%% file: main.tex
\documentclass[letterpaper]{article} 
\usepackage{aaai2027}  
\usepackage[hyphens]{url}  
\usepackage{graphicx} 
\usepackage{natbib}  
\usepackage{caption} 
\usepackage{algorithm}
\usepackage{algorithmic}

\usepackage{newfloat}
\usepackage{listings}
\DeclareCaptionStyle{ruled}{labelfont=normalfont,labelsep=colon,strut=off} 
\floatstyle{ruled}
\newfloat{listing}{tb}{lst}{}
\floatname{listing}{Listing}

\usepackage{booktabs}

\usepackage{amssymb}
\usepackage{makecell}
\usepackage{amsmath}
\usepackage{cleveref}
\usepackage{capt-of}
\usepackage{multirow}
\usepackage[table]{xcolor}

\title{2Xplat: Decoupling Geometry and Appearance Modeling \\ for Feed-Forward 3D Gaussian Splatting}
\author{
    Hwasik Jeong\textsuperscript{\rm 1}\equalcontrib,
    Seungryong Lee\textsuperscript{\rm 2}\equalcontrib,
    Gyeongjin Kang\textsuperscript{\rm 2},
    Seungkwon Yang\textsuperscript{\rm 1},
    Xiangyu Sun\textsuperscript{\rm 2},
    \\ Seungtae Nam\textsuperscript{\rm 1},
    Eunbyung Park\textsuperscript{\rm 1}\corresponding
}
\affiliations{
    \textsuperscript{\rm 1}Yonsei University\\
    \textsuperscript{\rm 2}Sungkyunkwan University\\
}

\makeatletter
\let\@oldmaketitle\@maketitle
\renewcommand{\@maketitle}{%
  \@oldmaketitle
  \vspace{-1mm}
  \centering
  \includegraphics[width=1.0\textwidth]{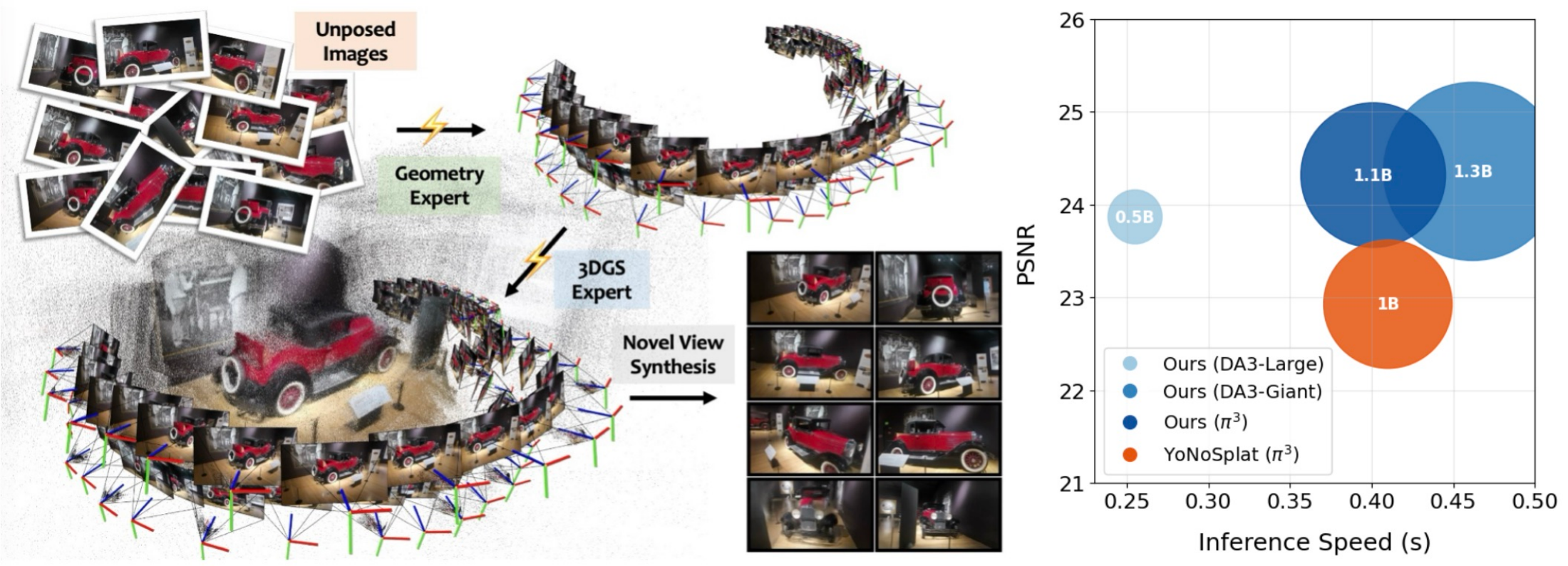}
    \captionof{figure}{Left: An illustration of our \textit{2Xplat} on a DL3DV scene. Right: PSNR vs. Inference speed (DL3DV, $224 \times 224$ resolution, 12 input views).}
  \label{fig:teaser}
  \vspace{3mm}
}
\makeatother

\begin{document}

\maketitle

\begin{abstract}
Pose-free feed-forward 3D Gaussian Splatting (3DGS) has opened a new frontier for rapid 3D modeling, enabling high-quality Gaussian representations to be generated from uncalibrated multi-view images in a single forward pass. The dominant approach adopts unified monolithic architectures, often built on geometry-centric 3D foundation models, to jointly estimate camera poses and synthesize 3DGS representations within a single network, entangling geometric reasoning and appearance modeling within a shared representation. In this work, we introduce \textit{2Xplat}, a pose-free feed-forward 3DGS framework based on a two-experts design that explicitly separates geometry estimation from Gaussian generation: a dedicated geometry expert first predicts camera poses, which are then passed to an appearance expert that synthesizes 3D Gaussians. Despite its conceptual simplicity, and being largely underexplored in prior works, our two-experts pipeline outperforms prior pose-free feed-forward 3DGS approaches in fewer than 5K training iterations, achieving performance on par with state-of-the-art posed methods. These results challenge the prevailing unified paradigm and suggest the potential advantages of modular design for complex 3D geometric estimation and appearance synthesis tasks.
\end{abstract}


\section{Introduction}

3D Gaussian Splatting (3DGS)~\cite{kerbl20233d} has recently emerged as a powerful representation for high-quality, real-time novel-view synthesis, enabling applications such as AR/XR~\cite{jiang2024vrgs, jiang2024dualgs}, immersive telepresence~\cite{gsnexus, telegs}, volumetric video production~\cite{sun2024splatter, shen2025nutshell}, robotics~\cite{escontrela2025gaussgym, lu2024manigaussian}, and autonomous driving~\cite{zhou2024drivinggaussian, hess2025splatad}. However, conventional 3DGS relies on computationally intensive per-scene optimization, often requiring tens of minutes to hours~\cite{kerbl20233d, yu2024mip}. Feed-forward 3DGS methods address this by directly predicting Gaussian parameters from multi-view images in a single pass~\cite{charatan2024pixelsplat, chen2024mvsplat, zhang2024gslrm, xu2025depthsplat}, reducing reconstruction time to seconds while matching optimization-based quality.

Despite these advances, many feed-forward approaches assume access to accurate camera poses, which limits their applicability in unconstrained settings where reliable pose estimates are unavailable~\cite{hartley2003multiple, schoenberger2016sfm, schoenberger2016mvs}, motivating pose-free feed-forward 3DGS methods~\cite{ye2025no, kang2025selfsplat, jiang2025anysplat, ye2026yonosplat} that reconstruct Gaussian representations directly from uncalibrated multi-view images.

Most existing pose-free feed-forward approaches adopt a monolithic design, in which a single network jointly predicts camera poses and 3DGS parameters using shared features with task-specific output heads~\cite{ye2025no, jiang2025anysplat, ye2026yonosplat}. We argue that this unified architecture may not be optimal for high-quality novel view synthesis: since appearance is also shaped by factors beyond geometry, such as lighting and view-dependent reflectance, strict geometric accuracy is not always sufficient for faithful appearance modeling, creating conflicting objectives and preventing full exploitation of pose-conditioned architectural mechanisms such as Epipolar Transformer~\cite{he2020epipolar}, PRoPE~\cite{li2025prope}, GTA~\cite{miyato2024gta}, CaPE~\cite{xiong2023cape}, and RayRoPE~\cite{wu2026rayrope} that have proven effective in posed feed-forward settings.

In this work, we adopt a two-stage design that combines a strong geometry estimator with a powerful, pose-conditioned 3DGS generator. This decoupling enables the appearance expert to fully leverage advanced pose-conditioned architectures, and by reusing two mature pretrained experts rather than introducing randomly initialized task-specific modules, the entire pipeline can be optimized efficiently through lightweight end-to-end fine-tuning, converging in fewer than 5K iterations. As a result, our approach consistently improves on prior pose-free feed-forward 3DGS methods while achieving competitive performance with state-of-the-art posed feed-forward methods.

Prior "geometry-first, appearance-second" approaches have explored similar decompositions~\cite{lai2021videoae, kang2025selfsplat, jiang2025rayzer}, but they primarily emphasize self-supervised training strategies rather than leveraging recent advances in high-capacity, pose-conditioned appearance models, leaving their novel-view synthesis quality behind state-of-the-art posed methods. In this work, we revisit this paradigm from a different perspective, explicitly combining a strong geometry estimator with a powerful pose-conditioned 3DGS generator to push pose-free novel-view synthesis quality to match or even surpass that of posed approaches. Our key contributions can be summarized as follows:
\begin{itemize}
    \item We explore an end-to-end two-experts framework that decomposes pose-free feed-forward 3DGS into a dedicated geometry expert and an appearance expert.
    \item Through lightweight end-to-end joint optimization, our appearance expert becomes robust to noisy camera pose estimates, mitigating the sensitivity of 3DGS generation to geometric errors.
    \item Our approach consistently improves on prior pose-free feed-forward 3DGS methods and performs on par with state-of-the-art posed models in novel view synthesis.
\end{itemize}

\begin{figure*}[t]
\centering
\includegraphics[width=\textwidth]{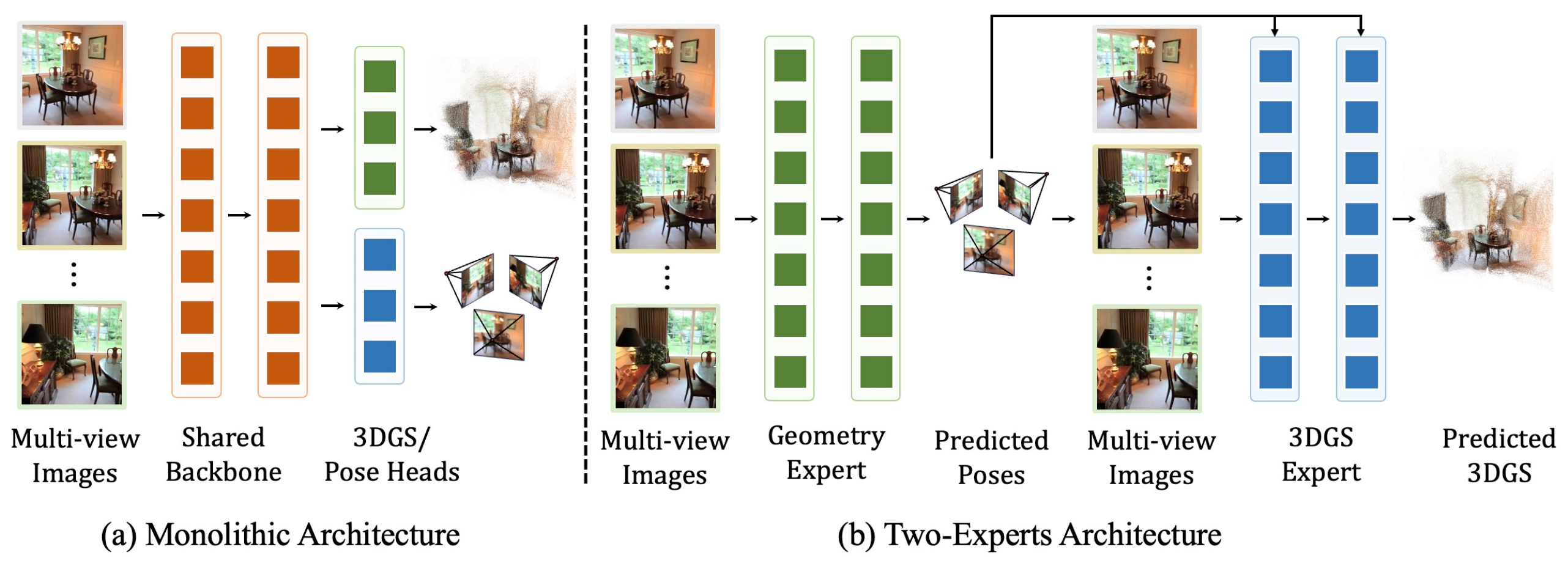}
\caption{
Two experts vs. one generalist: (a) The prevailing monolithic architecture employs a shared backbone with task-specific heads to jointly predict camera poses and 3DGS representations in a single forward pass. (b) The two-experts architecture explicitly decomposes the process into sequential stages: a geometry expert first estimates camera poses from multi-view input images, and a dedicated appearance (posed feed-forward 3DGS) expert subsequently generates 3D Gaussian representations conditioned on the predicted poses and input images.
}
\label{fig:two_experts_main}
\end{figure*}

\section{Related Works}

\subsection{Feed-forward 3D Foundation Models}
Traditional 3D reconstruction methods rely on per-scene optimization pipelines such as Structure-from-Motion~\cite{schoenberger2016sfm} and Multi-View Stereo~\cite{schoenberger2016mvs, hartley2003multiple}, which are computationally expensive and brittle to sparse or unstructured inputs, motivating a shift toward feed-forward approaches that amortize reconstruction cost through large-scale training. Building on Vision Transformers~\cite{dosovitskiy2020vit}, DUSt3R~\cite{wang2024dust3r} and MASt3R~\cite{leroy2024mast3r} pioneered this paradigm by framing pairwise reconstruction as dense pointmap regression, jointly predicting camera poses and geometry in a single forward pass; however, they operate primarily on image pairs, requiring a global alignment post-processing step to scale to multi-view inputs. More recent work removes this constraint by directly operating over arbitrary numbers of views~\cite{wang2025vggt, wang2026pi, lin2025depth3}, using cross-view attention to jointly reason about geometry and camera parameters while significantly reducing inference latency compared to alignment-based pipelines.

\subsection{Posed Feed-forward 3D Models}
A large body of feed-forward 3D reconstruction methods conditions on known camera poses at test time, offloading pose estimation to an external system such as SfM~\cite{schoenberger2016sfm}, with LRM~\cite{hong2024lrm} establishing this paradigm via a large-scale transformer mapping images to a neural radiance field~\cite{fridovich2023kplane} in a single pass. These methods fall into two categories by representation: explicit methods predict 3D primitives (e.g., Gaussians) using epipolar constraints~\cite{charatan2024pixelsplat}, cost-volume matching~\cite{chen2024mvsplat, xu2025depthsplat}, iterative refinement~\cite{nam2025generative, kang2026ilrm}, or end-to-end regression~\cite{zhang2024gslrm, imtiaz2025lvt, kang2026multi}, while implicit methods instead train transformers to directly render novel views from posed images~\cite{flynn2024quark, jin2024lvsm}. While these methods achieve impressive reconstruction quality and fast inference, they fundamentally assume accurate camera poses are available at test time.

\subsection{Pose-free Feed-forward 3D Models}
To remove the dependency on known camera poses, a growing line of work explores feed-forward 3D reconstruction from unposed images, jointly inferring scene geometry, appearance, and camera parameters in a single pass, spanning representations from neural fields~\cite{jiang2024leap} to 3D Gaussian Splatting~\cite{ye2025no, kang2025selfsplat, ye2026yonosplat, jiang2025anysplat, sun2026uni3r}, and demonstrating that accurate geometry and photorealistic appearance can be recovered without any pose inputs at inference time. Despite this progress, prevailing pipelines~\cite{zhang2025flare, ye2025no, jiang2025anysplat, ye2026yonosplat, sun2026uni3r} share a common bottleneck: a single monolithic network jointly estimates camera poses and Gaussian parameters from shared features, entangling two distinct objectives and imposing an inherent performance ceiling on both. FLARE~\cite{zhang2025flare} is most closely related to our work, as both use camera pose as an explicit bridge between geometry and appearance; however, FLARE trains all components jointly from scratch in a monolithic manner, whereas our two-expert framework starts from independently pretrained geometry and appearance experts, combined via lightweight end-to-end fine-tuning so that each objective is handled by a specialized module while remaining tightly coupled through joint optimization.

\section{Method}

\subsection{Problem Formulation}
We consider the pose-free feed-forward 3DGS task, where the goal is to generate a 3D Gaussian representation directly from unposed multi-view images. Optionally, the model may also estimate the camera pose associated with each input view. Formally, let $\{ I_i \in \mathbb{R}^{H \times W \times 3} \}_{i=1}^N$ denotes a set of $N$ input images with height $H$ and width $W$. A pose-free feed-forward model $F(\cdot)$ maps input images to a set of pixel-aligned 3D Gaussians and camera parameters.
\begin{equation}
    \{\hat{p}_i \}_{i=1}^N, \{G_j\}_{j=1}^{N_c} = F(\{ I_i \}_{i=1}^N),
\end{equation}
where $N=N_c+N_t$, and $N_c$ and $N_t$ denote the numbers of context and target views, respectively. $G_j \in \mathbb{R}^{H \times W \times d_g}$ denotes pixel-aligned 3D Gaussian for each context view image. The camera parameters $p_i=[K_i, R_i, t_i] \in \mathbb{R}^{d_p}$ represent the intrinsic $K_i$ and extrinsic components $(R_i, t_i)$ corresponding to the image $I_i$, and $ d_g$ and $d_p$ are the dimensions of the 3D Gaussian attributes and camera parameters, respectively.

\subsection{Monolithic vs. Two-Experts Architecture}
A common architectural paradigm for pose-free feed-forward 3DGS adopts a monolithic design, where a single network jointly predicts camera poses and 3D Gaussian parameters using a shared backbone with task-specific output heads; as discussed in the introduction, this may entangle geometry estimation and appearance modeling within a shared representation, limiting specialization, hindering the exploitation of recent pose-conditioned architectural designs, and lacking the capacity required for high-fidelity 3DGS generation. Recent approaches further extend this paradigm by allowing camera poses to be optionally provided as input, enabling a single model to handle both posed and pose-free settings~\cite{lin2025depth3, ye2026yonosplat}; however, this unified formulation requires the network to implicitly switch between predicting poses when unavailable and bypassing pose prediction when they are provided. Learning this behavior within a shared representation can be non-trivial and may complicate the integration of advanced pose-conditioned architectural mechanisms.

We instead explore a two-experts framework that explicitly decomposes geometry estimation and 3DGS generation into sequential modules: a pose expert $F_{\text{pose}}$ estimates camera parameters from input images, and an appearance expert $F_{\text{3dgs}}$ generates pixel-aligned 3D Gaussian representations conditioned on the context view images and predicted poses.
The entire pipeline remains end-to-end trainable, enabling the 3DGS generator to become robust to pose estimation errors through joint optimization, while allowing the pose expert to be simply bypassed when ground-truth camera parameters are available, so the appearance expert can directly operate in the posed setting. This modular design naturally accommodates both scenarios and enables independent incorporation of architectural advancements from both geometry estimation and posed feed-forward 3DGS. While separating the pose and appearance modules raises concerns about redundant low-level visual processing, our empirical results show that this does not compromise efficiency and, in fact, proves beneficial. With comparable, and in some cases fewer parameters than monolithic counterparts (\cref{tab:size_comparison}), our two-experts framework consistently achieves significantly better performance (\cref{tab:monolithic_vs_twoexperts}), though exploring more principled sharing of low-level reasoning between the two experts remains an interesting direction for future work.

\subsection{Geometry Expert}
Recent advances in large-scale 3D geometry foundation models such as DUSt3R~\cite{wang2024dust3r}, VGGT~\cite{wang2025vggt}, $\pi^3$~\cite{wang2026pi}, and Depth Anything 3 (DA3)~\cite{lin2025depth3} have significantly improved multi-view geometry estimation, trained on extensive synthetic and real-world datasets with dense depth, point map, ray map, and camera pose annotations. In particular, DA3 demonstrates that training with depth, pointmap, ray map, and auxiliary camera pose objectives yields state-of-the-art results in both pose accuracy and geometry reconstruction, and we thus adopt DA3 as our geometry expert. For a fair comparison with prior pose-free feed-forward 3DGS methods~\cite{ye2025no, ye2026yonosplat}, we additionally evaluate alternative geometry backbones to ensure that the benefits of our two-experts framework are not tied to a single geometry model (\cref{tab:monolithic_vs_twoexperts}).

\subsection{Appearance Expert}
For the 3DGS expert, we adopt the recent Multi-view Pyramid Transformer (MVP) architecture~\cite{kang2026multi}, which achieves strong performance among posed feed-forward 3DGS models in terms of both reconstruction quality and inference efficiency. MVP integrates PRoPE-based camera pose conditioning~\cite{li2025prope}, register tokens~\cite{darcet2024vision}, and an Alternating Attention design~\cite{wang2025vggt} within a dual hierarchical framework, trained entirely from scratch without relying on pretrained DINO features~\cite{oquab2023dinov2, simeoni2025dinov3}. Its computational efficiency enables finer-grained spatial tokens, improving reconstruction fidelity without prohibitive cost. However, our framework is not tied to this specific choice and remains compatible with other posed feed-forward 3DGS models, as we demonstrate in \cref{tab:appearance_expert_ablation}.

While our appearance model performs well overall, its global attention lacks 3D-consistency, which can lead to unrealistic geometry in the generated 3DGS. We hypothesize that a more 3D-consistent model would improve generalization performance, and based on this insight, we introduce a more powerful, 3D-consistent variant that combines a dynamic grouping mechanism with a larger network capacity and a substantially larger training set of approximately 34M samples, while keeping training cost equivalent to that of the original model. We provide detailed results and analysis of this variant in the Appendix, as the comparison is not entirely fair due to differences in dataset size and model capacity relative to the baselines.

Monolithic feed-forward 3DGS models~\cite{ye2026yonosplat, lin2025depth3} typically retain the $14 \times 14$ (or $16 \times 16$) patch size of their pretrained backbones, as changing it breaks compatibility with the pretrained weights. In our two-expert framework, geometry and appearance are decoupled, and the appearance expert does not rely on a large pretrained backbone. This allows us to use a smaller patch size for the appearance expert, e.g., $4 \times 4$, while keeping the geometry expert unchanged, to better preserve fine-grained local details.

\input{Tables/dl3dv_nvs}

\begin{figure*}[!h]
\centering
\includegraphics[width=\textwidth]{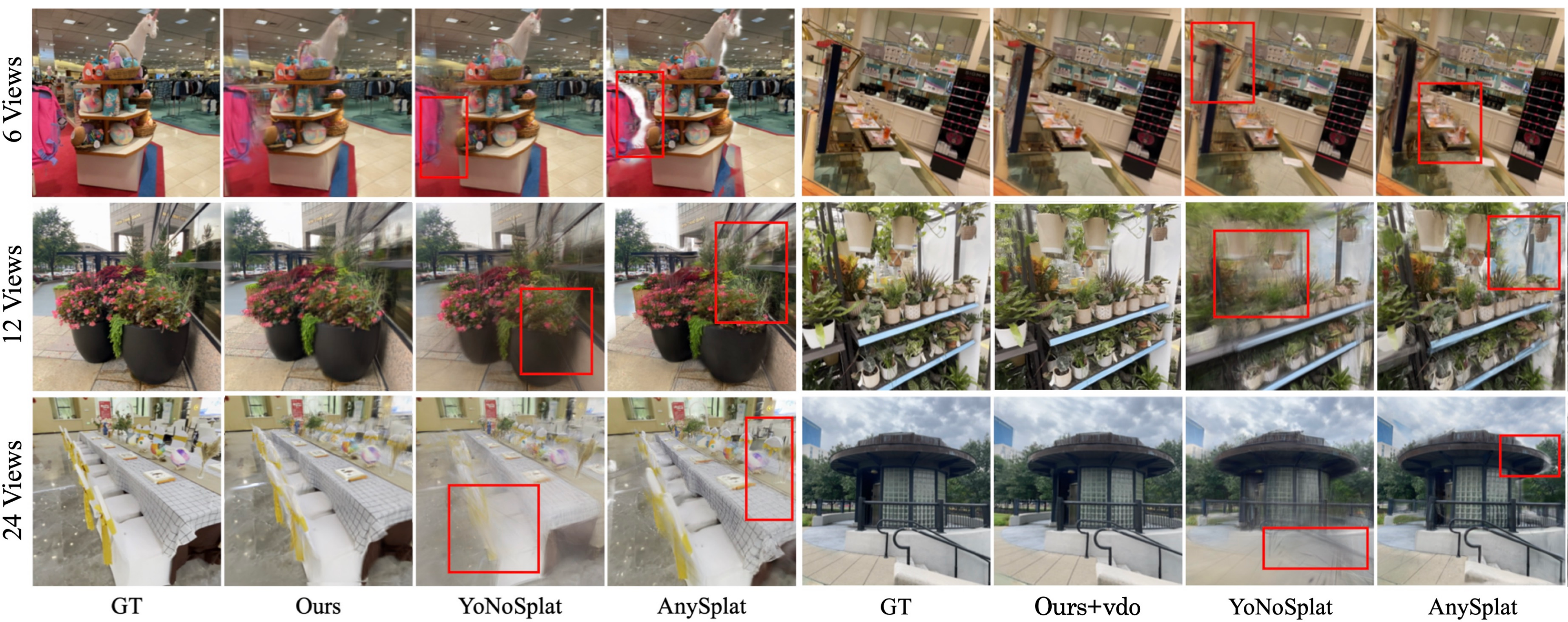}
\vspace{-0.5cm}
\caption{
Qualitative comparison on DL3DV with varying numbers of input views.
}
\label{fig:dl3dv_lr}
\end{figure*}

\subsection{Joint Training}
We initialize our framework from two pretrained experts, a geometry (camera pose) expert and an appearance (3DGS) expert, and jointly train the entire system by fine-tuning both experts end-to-end. Given a set of context and target views ${ \{I_i\} }_{i=1}^{N}$, where $N=N_c + N_t$, the pose expert first predicts camera parameters for all views,
\begin{equation}
\{\hat{p}_i \}_{i=1}^{N} = F_\text{pose}(\{ I_i \}_{i=1}^{N}).
\end{equation}
The 3DGS expert then takes the context images together with their predicted camera parameters to generate pixel-aligned 3D Gaussian representations,
\begin{equation}
\{G_i\}_{i=1}^{N_c} = F_\text{3dgs}(\{ I_i \}_{i=1}^{N_c}, \{\hat{p}_i \}_{i=1}^{N_c}).
\end{equation}
Using a differentiable 3DGS renderer, we render each target view from the predicted Gaussian representation and compute an image reconstruction loss against the corresponding ground-truth image. To regularize pose prediction and prevent geometric drift, we additionally supervise the predicted poses with ground-truth camera parameters. The overall training objective $\mathcal{L}$ is defined as
\begin{equation}
\mathcal{L} = \frac{1}{N_t}\sum_{i=1}^{N_t}\mathcal{L}_\text{render}(\hat{I}_i, I_{N_c+i})+ \frac{1}{N}\sum_{j=1}^N\mathcal{L}_\text{cam}(\hat{p}_j, p_j).
\end{equation}
Here, $\hat{I}_i$ and $I_{N_c+i}$ denote the rendered target view and its corresponding ground-truth image, while $\hat{p}_j$ and $p_j$ are the predicted and ground-truth camera parameters. The rendering loss $\mathcal{L}_{\text{render}}$ consists of a weighted combination of $\ell_2$ reconstruction loss and perceptual loss, balancing pixel-wise accuracy and perceptual fidelity. The camera supervision term $\mathcal{L}_{\text{cam}}$ consists of a relative pose loss following~\cite{wang2026pi, ye2026yonosplat}, which addresses the ambiguity in the global reference frame between the predicted and ground-truth poses. 

Optionally, when the appearance expert employs view-dependent opacity~\cite{imtiaz2025lvt}, we further regularize the viewpoint-dependent splat opacities to suppress opacity artifacts that are inconsistent across viewpoints. Let $N_G$ denote the total number of predicted Gaussian primitives (in the per-pixel Gaussian setting, $N_G = N \cdot H \cdot W$). The regularization term $R_\alpha$ is defined as
\begin{equation}
R_\alpha = \frac{1}{N_G} \sum_{j=1}^{N_G} \left| \sigma(\alpha_j \cdot \omega_j) \right|,
\label{eq:reg_alpha}
\end{equation}
where $\sigma(\cdot)$ is the Sigmoid function and $\omega_j$ is the spherical harmonic basis evaluated at a randomly sampled per-pixel viewing direction. This term is added to the total loss with a weight of 0.001. Full loss formulations are provided in the Appendix.

\input{Tables/dl3dv_hr}

\begin{figure*}[h]
\centering
\includegraphics[width=\textwidth]{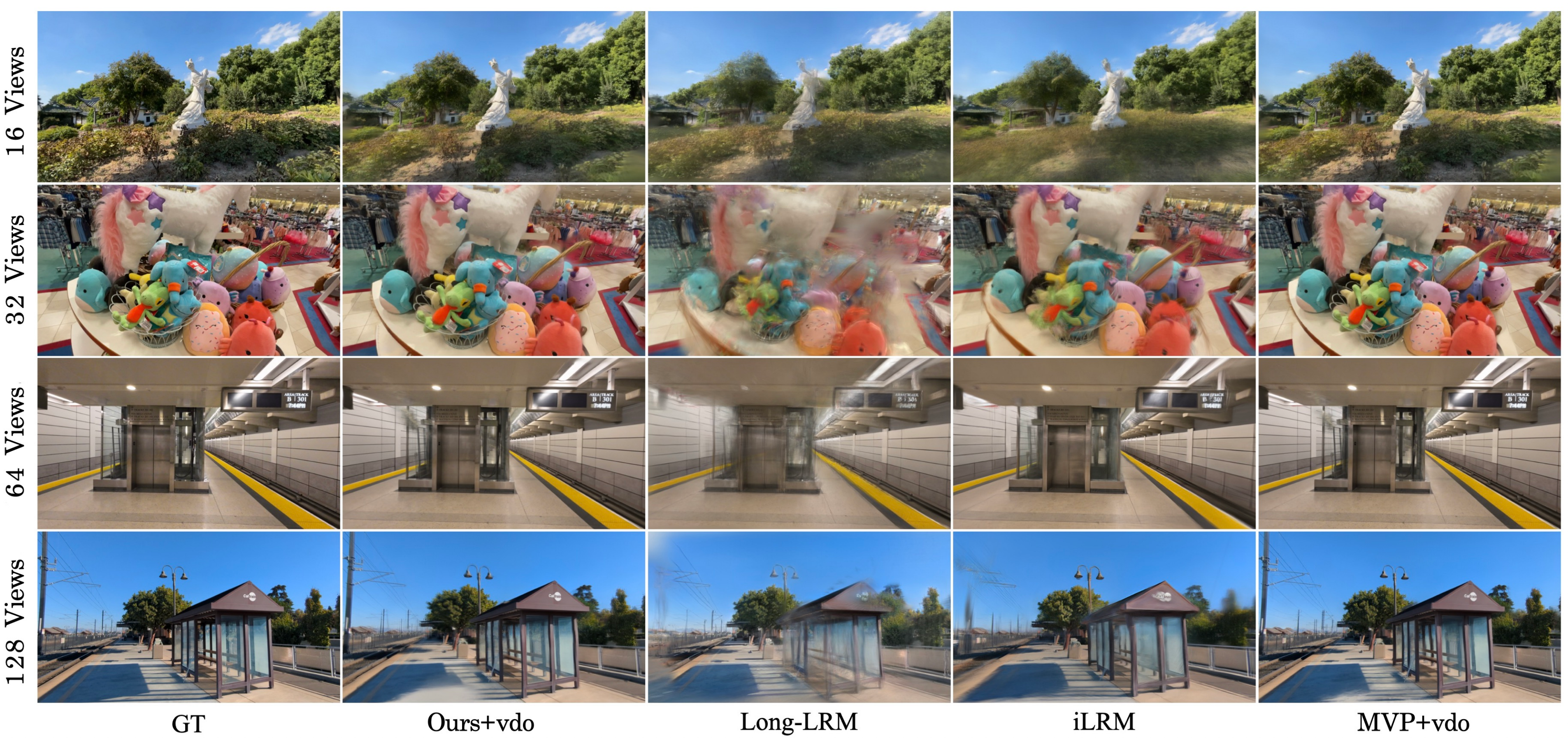}
\caption{Qualitative comparison on high-resolution DL3DV ($960 \times 540$).}
\label{fig:dl3dv_hr}
\end{figure*}

\section{Experiments}

\noindent{\textbf{Dataset.}} We train our model on the RealEstate10K (RE10K)~\cite{zhou2018stereo} and DL3DV~\cite{ling2024dl3dv} datasets using their official data splits. For benchmarking on RE10K, we retain only test sequences with at least 200 frames, yielding 1,580 sequences, and follow prior work~\cite{ye2026yonosplat} in using 6 context views for fair comparison. For DL3DV, we evaluate performance using 6, 12, and 24 input views, with maximum frame intervals of 50, 100, and 150, respectively; input views are selected via farthest point sampling over camera centers, with 8 views randomly held out for validation, following~\cite{ye2026yonosplat}. For the high-resolution DL3DV evaluation, we use the undistorted version of the dataset following~\cite{ziwen2025long, kang2026multi} and adopt the same evaluation protocol. To assess low-resolution generalization, we evaluate the DL3DV-trained model on ScanNet++~\cite{yeshwanth2023scannet++}, sampling 32, 64, and 128 views per scene with a fixed target view.

\noindent\textbf{Baselines.}
For novel view synthesis, we compare with pose-dependent methods (MVSplat~\cite{chen2024mvsplat}, DepthSplat~\cite{xu2025depthsplat}, Long-LRM~\cite{ziwen2025long}, iLRM~\cite{kang2026ilrm}, MVP~\cite{kang2026multi}) and pose-free methods (NoPoSplat~\cite{ye2025no}, AnySplat~\cite{jiang2025anysplat}, YonoSplat~\cite{ye2026yonosplat}).

\begin{figure}[h!]
\centering
\includegraphics[width=\columnwidth]{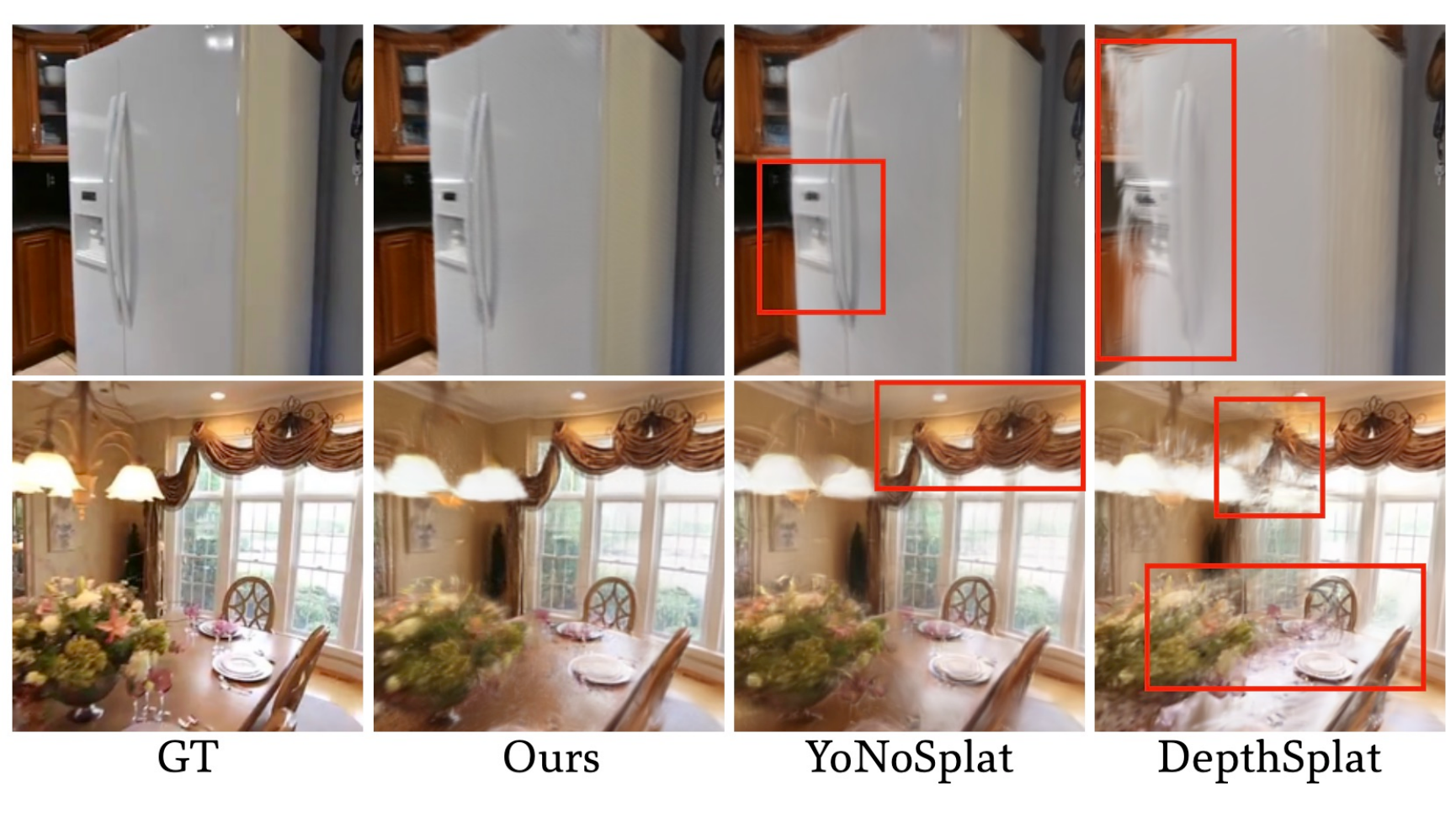}
\caption{Qualitative comparison on low-resolution RE10K dataset with 6 context views.}
\label{fig:re10k_comparison}
\end{figure}

\input{Tables/re10k_nvs}

\noindent\textbf{Evaluation Protocol.} For novel view synthesis, we adopt standard image quality metrics: PSNR, SSIM~\cite{ssim}, and LPIPS~\cite{zhang2018lpips}. We report results under both pose-dependent and pose-free protocols, rendering with ground-truth poses in the former and predicted poses in the latter; since several prior pose-free methods~\cite{ye2025no, ye2026yonosplat} adopt evaluation-time pose alignment (EPA), we additionally report EPA results, marked with $^{\dagger}$, for fair comparison. Unless otherwise specified, all results are evaluated at low resolution and without view-dependent opacity (vdo).

\input{Tables/dl3dv_2_scannetpp_nvs}

\begin{figure}[!h]
\centering
\includegraphics[width=0.48\textwidth]{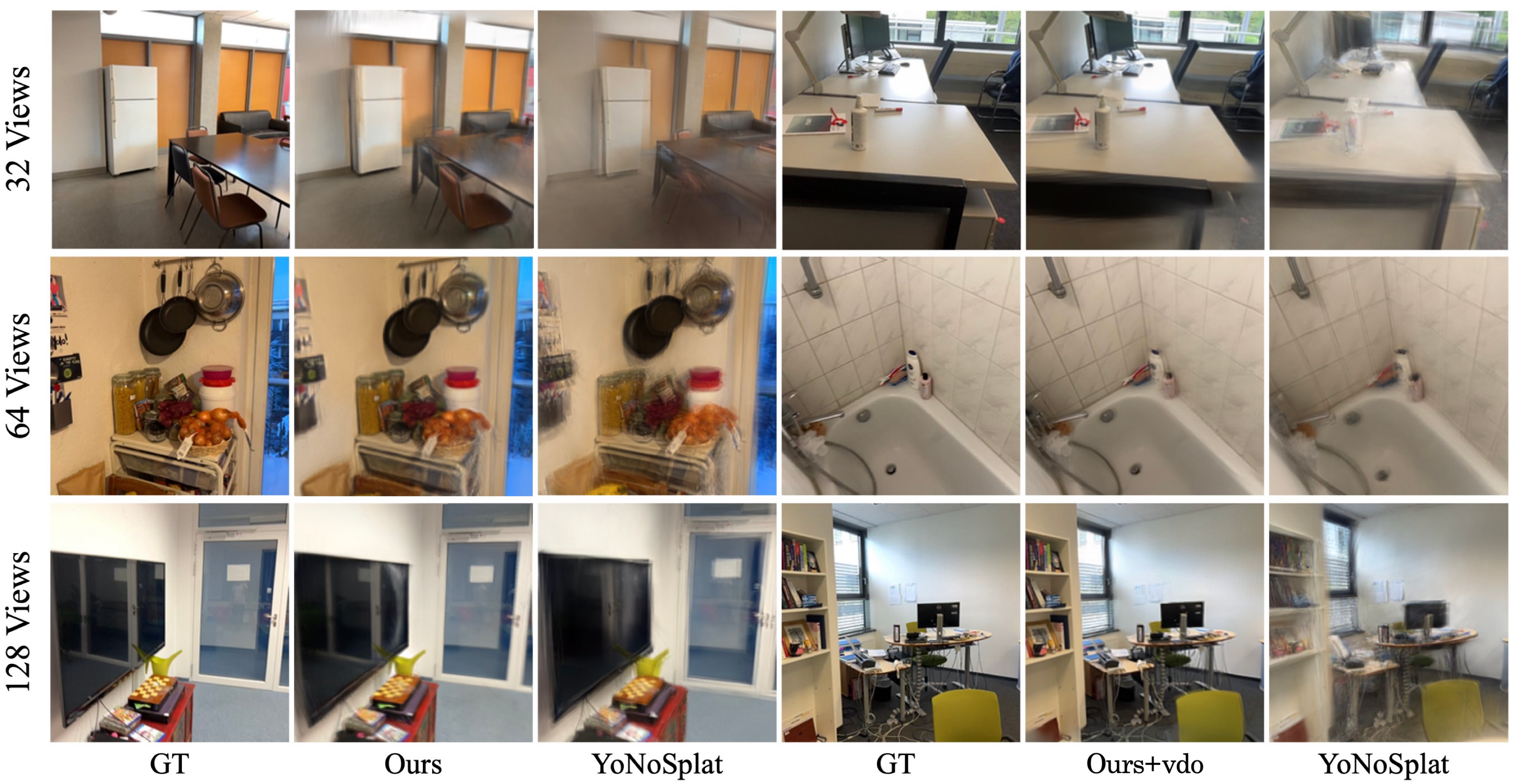}
\caption{Qualitative cross-dataset generalization from DL3DV to ScanNet++.}
\label{fig:scan_lr}
\end{figure}

\section{Results}

\noindent\textbf{Novel View Synthesis.} We evaluate our method in a low-resolution setting on both DL3DV (outdoor) and RE10K (indoor). On DL3DV, we vary the number of input views, while on RE10K we evaluate under a fixed-view setting. As shown in \cref{tab:nvs_dl3dv} and \cref{tab:re10k_nvs}, our pose-free and intrinsic-free model consistently outperforms all baselines on both datasets, including those that rely on ground-truth camera poses or intrinsics, demonstrating the effectiveness of our two-expert framework for pose-free novel view synthesis across both indoor and outdoor scenarios. On DL3DV, this performance gap widens consistently as the number of input views increases, while competing methods degrade noticeably under the same conditions, showing that our approach is substantially more robust to the number of input views. Notably, prior methods such as YoNoSplat~\cite{ye2026yonosplat} and NoPoSplat~\cite{ye2025no} depend heavily on evaluation-time pose alignment (EPA) to reach competitive performance, whereas our method already surpasses these baselines without EPA and improves further when EPA is applied. Qualitative comparisons on DL3DV and RE10K are shown in \cref{fig:dl3dv_lr} and \cref{fig:re10k_comparison}, respectively, where our method produces sharper structures and more consistent renderings than prior approaches on both datasets.

We further evaluate our method on the DL3DV dataset in a high-resolution setting (\cref{tab:quantitative_result_on_dl3dv_hr}, \cref{fig:dl3dv_hr}). We compare against the optimization-based 3D Gaussian Splatting~\cite{kerbl20233d} (30K iterations) and feed-forward reconstruction methods~\cite{ziwen2025long, kang2026ilrm, kang2026multi}. Across all settings from 16 to 128 input views, ours is the only pose-free method. Nevertheless, our method achieves competitive performance.

\noindent\textbf{Cross-Dataset Generalization.}
To evaluate cross-dataset generalization, we train our model on DL3DV and directly test it on ScanNet++ at low resolution, without any fine-tuning. As shown in \cref{tab:dl3dv_2_scannetpp}, our method consistently outperforms the baselines across all evaluation metrics in this cross-dataset setting, demonstrating that our two-expert framework generalizes well across different scene distributions. Qualitative comparisons in \cref{fig:scan_lr} further confirm this, with our method producing sharper and more view-consistent renderings than the baselines.

\input{Tables/dl3dv_sparse_view}

\noindent\textbf{Sparse-View Evaluation.} Since pose-free methods typically struggle in sparse-view settings, we additionally evaluate our model with only 2 context views. For a fair comparison, we lightly fine-tune our appearance model for 50K iterations with 2 context views, whereas baseline models are extensively trained specifically for sparse-view settings. As shown in \cref{tab:sparse_view_ablation}, our model remains competitive with these baselines despite requiring substantially less fine-tuning.

\section{Analyses}

\noindent\textbf{Monolithic vs. Two-Experts Architectures.} To demonstrate that a two-experts design outperforms a monolithic design, we compare our two-experts framework against three monolithic architectures: DA3-G (DA3 Giant), DA3-L (DA3 Large), and YoNoSplat. We evaluated DA3-G following the protocol in \cite{lin2025depth3}. Since DA3-L lacks a Gaussian head, we used it from DA3-G and trained the model for 150K iterations using RGB and camera losses.
During this training, the 3D Gaussians' positions are projected using the predicted depth and raymap, while the remaining attributes are produced by the Gaussian head.
For a fair comparison, both the monolithic baselines and our framework use the same backbones.
As shown in \cref{tab:monolithic_vs_twoexperts}, our framework consistently outperforms the monolithic architectures across all input view counts.

\input{Tables/monolithic_vs_twoexperts}

\noindent\textbf{Robustness to Geometry Experts.}
We evaluate three geometry backbone choices for camera pose estimation: DA3-L, DA3-G, and $\pi^3$, as shown in \cref{tab:size_comparison} and \cref{tab:monolithic_vs_twoexperts}. In \cref{tab:size_comparison}, Ours paired with each backbone (DA3-G, DA3-L, $\pi^3$) denotes our framework using that backbone as the geometry expert. Our method performs consistently well across all backbone choices, demonstrating its robustness. Notably, even with DA3-L, which has fewer parameters, our method still achieves competitive performance with faster inference speed, indicating that our approach does not rely heavily on backbone capacity. Furthermore, replacing the backbone with larger variants such as DA3-G leads to only moderate improvements, suggesting that the performance gain mainly stems from our architecture rather than increased model size. This highlights the efficiency of our design in effectively leveraging the geometry expert.

\input{Tables/joint_training_ablation}

\noindent\textbf{Joint Training Analysis.} To assess the impact of joint training in our framework, we compare models trained with and without it. As shown in \cref{tab:joint_training_ablation}, jointly training the two experts yields a substantial performance improvement, confirming that joint optimization is essential to our framework's effectiveness.

\section{Conclusion}

In this work, we present \textit{2Xplat}, a two-experts framework for pose-free feed-forward 3DGS that decouples pose estimation from appearance synthesis. Despite its conceptual simplicity, our approach outperforms prior pose-free methods and achieves performance on par with state-of-the-art posed approaches, all within fewer than 5K training iterations, demonstrating strong reconstruction quality with remarkable training efficiency. These findings challenge the assumption that entangling geometric reasoning and appearance modeling within a shared representation is necessary or beneficial, and instead highlight the potential of modular design principles for complex 3D reconstruction tasks. We hope this work motivates further exploration of expert-decomposed architectures in 3D generation and beyond.

\begin{center}
\large
\textbf{Appendix}
\end{center}

\section{Additional Details}

\noindent\textbf{Joint Training Loss Formulation.} As introduced in the main text, the overall training objective $\mathcal{L}$ is defined as
\begin{equation}
\mathcal{L} = \frac{1}{N_t}\sum_{i=1}^{N_t}\mathcal{L}_\text{render}(\hat{I}_i, I_{N_c+i})+ \frac{1}{N}\sum_{j=1}^N\mathcal{L}_\text{cam}(\hat{p}_j, p_j).
\end{equation}
Here, $\hat{I}_i$ and $I_{N_c+i}$ denote the rendered target view and its corresponding ground-truth image, while $\hat{p}_j$ and $p_j$ are the predicted and ground-truth camera parameters. For the rendering loss $\mathcal{L}_{\text{render}}$, we adopt a weighted combination of $\ell_2$ reconstruction loss and perceptual loss to balance pixel-wise accuracy and perceptual fidelity,
\begin{equation}
\mathcal{L}_\text{render} = \mathcal{L}_{MSE}(\hat{I}, I) + \lambda_{perc}\mathcal{L}_{perc}(\hat{I}, I)
\end{equation}
where $\lambda_{perc}$ controls the contribution of the perceptual term. For the camera supervision term $\mathcal{L}_{\text{cam}}$, we adopt a relative pose loss following ~\cite{wang2026pi, ye2026yonosplat} to address the ambiguity in the global reference frame between the predicted and ground-truth poses. Specifically, the predicted relative pose $\hat{T}_{i\leftarrow j}$ from view $j$ to $i$ is computed as
\begin{equation}
\hat{T}_{i \leftarrow j} 
= \hat{T}_i^{-1}\hat{T}_j
=
\begin{bmatrix}
\hat{R}_{i \leftarrow j} & \hat{t}_{i \leftarrow j} \\
0 & 1
\end{bmatrix}
\end{equation}
where $\hat{T}_{i \leftarrow j}$ consists of the relative rotation $\hat{R}_{i \leftarrow j}$ and translation $\hat{t}_{i \leftarrow j}$. The relative rotation loss $L_R(i,j)$ and translation loss $L_t(i,j)$ are defined as
\begin{equation}
\begin{split}
\mathcal{L}_R(i, j) 
&= 
\arccos\left(
\frac{
\operatorname{tr}
\left(
R_{i \leftarrow j}^{\top}
\hat{R}_{i \leftarrow j}
\right)
- 1
}{2}
\right), 
\\
\mathcal{L}_t(i, j) 
&= 
H_{\delta}
\left(
\hat{t}_{i \leftarrow j} - t_{i \leftarrow j}
\right)
\end{split}
\end{equation}
where $\operatorname{tr}(\cdot)$ denotes the trace of a matrix and $H_{\delta}(\cdot)$ represents the Huber loss with threshold $\delta$. The overall pose loss is the formulated as
\begin{equation}
\begin{split}
\mathcal{L}_{\text{cam}} 
&= 
\frac{1}{N(N - 1)} 
\sum_{i \ne j}
\left(
\lambda_R \mathcal{L}_R(i, j) + \lambda_t \mathcal{L}_t(i, j)
\right) \\
&\quad+
\frac{\lambda_K}{N} 
\sum_{j=1}^N
\mathcal{L}_K(j)
\end{split}
\end{equation}
where $\mathcal{L}_K$ denotes $l_2$ loss between the predicted and ground-truth camera intrinsics, and $\lambda_R$, $\lambda_t$ and $\lambda_K$ are weighting factors that balance between the contributions of each component. Optionally, we add view-dependent opacity regularization $R_{\alpha}$ when necessary, as described in the main text.

\noindent\textbf{Dataset.} For the DL3DV~\cite{ling2024dl3dv} and RealEstate10K (RE10K)~\cite{zhou2018stereo} datasets, we use the same low- and high-resolution splits described in the main text. To further evaluate high-resolution ($960 \times 540$) generalization, we test on three additional benchmarks, following the same protocol as \cite{kang2026multi}, the \texttt{train} and \texttt{truck} scenes from Tanks\&Temples~\cite{Knapitsch2017tnt}, nine scenes from Mip-NeRF360~\cite{barron2022mipnerf360} (\texttt{bicycle}, \texttt{bonsai}, \texttt{counter}, \texttt{garden}, \texttt{kitchen}, \texttt{room}, \texttt{stump}, \texttt{flower}, and \texttt{treehill}), and several scenes from ETH3D~\cite{schops2017multi}. To ensure a fair comparison across methods, all images are resized and cropped.

\noindent\textbf{Baselines} For novel view synthesis, we compare with both pose-dependent and pose-free methods, as described in the main text. For pose estimation, we compare with MASt3R~\cite{leroy2024mast3r}, VGGT~\cite{wang2025vggt}, $\pi^3$~\cite{wang2026pi}, and DA3~\cite{lin2025depth3}.

\noindent\textbf{Evaluation Protocol.} For novel view synthesis, we adopt PSNR, SSIM~\cite{ssim}, and LPIPS~\cite{zhang2018lpips}, as described in the main text. For pose estimation, we report the cumulative angular pose error curve (AUC) evaluated at thresholds of 5$^\circ$, 10$^\circ$, and 20$^\circ$~\cite{edstedt2024roma}. All other evaluation protocols follow the main text.

\noindent\textbf{Implementation Details.} We use the pretrained Depth Anything 3~\cite{lin2025depth3} and Multi-view Pyramid Transformer~\cite{kang2026multi} as the geometry and appearance experts, respectively, and adopt DA3-Giant unless otherwise specified. All input images are resized such that the shorter side matches the target resolution, followed by a center square crop. We use the same image resolution as~\cite{ye2026yonosplat} to ensure a fair comparison. For RE10K~\cite{zhou2018stereo}, we train at $224 \times 224$ resolution using 6 context views and 8 target views, with a batch size of 8 per GPU. For DL3DV~\cite{ling2024dl3dv} at $224\times224$ resolution, we use 6, 12, 24, and 32 input views with corresponding target views of 6, 6, 6, and 1, respectively, and per-GPU batch sizes of 4, 2, 1, and 1. For the high-resolution DL3DV setting, we use $960 \times 540$ images for the appearance expert and $504 \times 280$ for the geometry expert, training with 16 and 32 context views and corresponding target views of 4 and 1, with per-GPU batch sizes of 2 and 1.

All models are trained on 8 H200 GPUs within 5K iterations, whereas YoNoSplat~\cite{ye2026yonosplat} requires 16 GH200 GPUs and 150K iterations, highlighting the superior computational efficiency of our approach. During fine-tuning (referred to as joint training in our work), the model is optimized using a combination of rendering and relative pose losses, where the rendering loss includes a perceptual term with weight $\lambda_{\text{perc}} = 0.5$, and the relative pose loss uses weights $\lambda_R = 0.1$, $\lambda_t = 10$, and $\lambda_K = 0.5$ for rotation, translation, and intrinsic parameters, respectively. We train using the AdamW~\cite{loshchilov2017decoupled} optimizer with a learning rate of $2\times10^{-5}$, $\beta_1 = 0.9$, $\beta_2 = 0.99$, and a weight decay of $0.05$, applying gradient-norm-based clipping to stabilize end-to-end training. For evaluation-time pose alignment (EPA), we further refine all camera parameters for 100 iterations using the Adam~\cite{kingma2014adam} optimizer with a learning rate of $1\times10^{-4}$, following YoNoSplat~\cite{ye2026yonosplat} for fair comparison. To address the scale ambiguity between poses predicted by the geometry expert and ground-truth poses obtained from SfM, we normalize both by dividing translations by the maximum translation magnitude.

\section{Additional Analyses}

\noindent\textbf{Generality Across Appearance Experts.} To demonstrate that our two-experts framework generalizes across different appearance experts, we test it with several choices, including iLRM and Long-LRM, on the DL3DV high-resolution setting (\cref{tab:appearance_expert_ablation}). Compared to MVP, these experts degrade by nearly 1–2 PSNR due to their limited capacity to handle pose noise, yet they still perform reasonably well in a pose-free setting. All variants are jointly trained within 5K iterations, demonstrating the generality of the proposed two-experts framework across different appearance backbones.

\input{Tables/appearance_expert_ablation}

\noindent\textbf{Pose Estimation.} Although our primary focus lies in learning a high-quality 3DGS representation, our framework also enables accurate camera pose estimation as a byproduct. Notably, our method achieves competitive AUC performance despite being fine-tuned on only a small subset of the dataset, whereas prior state-of-the-art approaches~\cite{ye2026yonosplat, ye2025no} rely on extensive pose supervision (\allowbreak\cref{tab:pose_estimation}). This suggests that jointly leveraging complementary geometric and appearance modeling yields reliable pose estimates without requiring large-scale pose-specific training.

\begin{table}[h]
\centering
\small
\setlength{\tabcolsep}{6pt}
\begin{tabular}{l|ccc|c}
\toprule
Method & 5$^\circ\uparrow$ & 10$^\circ\uparrow$ & 20$^\circ\uparrow$ & Backbone \\
\midrule
MASt3R $_{518\times 288}$ & 0.609 & 0.776 & 0.878 & - \\
VGGT $_{518\times280}$ & 0.566 &  0.753 & 0.867 & - \\
$\pi^3$ $_{518\times280}$ & 0.705 &  0.841 & 0.916 & - \\
DA3 $_{504\times504}$ & 0.694 & 0.826 & 0.900 & - \\
\midrule
NoPoSplat$_{256\times 256}$ & 0.443 & 0.627 & 0.755 & MASt3R \\
YoNoSplat$_{224\times 224}$ & \textbf{0.722} & \textbf{0.852} & \textbf{0.923} & $\pi^3$  \\
Ours$_{224\times 224}$ & \underline{0.718} & \underline{0.843} & \underline{0.912} & DA3 \\
\bottomrule
\end{tabular}
\caption{Pose estimation comparison on RE10K $224 \times 224$.}
\label{tab:pose_estimation}
\end{table}

\noindent\textbf{Pose Supervision.}
We compare three pose loss configurations in~\cref{tab:pose_supervision}: relative loss, absolute loss, and without pose loss. While training without pose loss achieves marginally better rendering quality, it significantly degrades pose estimation accuracy across all angular thresholds. Therefore, we adopt relative loss, which achieves the best balance between rendering quality and pose accuracy.

\input{Tables/mvpv2_generalization_nvs}

\begin{table}[h]
\centering
\small
\setlength{\tabcolsep}{6pt}
\begin{tabular}{l|cc|ccc}
\toprule
Loss type & PSNR$\uparrow$ & LPIPS$\downarrow$ & 5$^\circ\uparrow$ & 10$^\circ\uparrow$ & 20$^\circ\uparrow$ \\
\midrule
Rel. loss & 26.161 & 0.131 & 0.718 & 0.843 & 0.912 \\
Abs. loss & 25.704 & 0.137 & 0.641 & 0.797 & 0.888 \\
W/O loss & 26.369 & 0.129 & 0.686 & 0.836 & 0.905 \\
\bottomrule
\end{tabular}
\caption{Pose supervision ablation on DL3DV $224 \times 224$.}
\label{tab:pose_supervision}
\end{table}

\begin{figure}[t]
\centering
\includegraphics[width=0.45\textwidth]{Figures/dynamic_grouping.pdf}
\caption{
Comparison between fixed and dynamic grouping. (a) Fixed grouping restricts each token to attend only within its assigned group, limiting network capacity even during later global attention. (b) Dynamic grouping relaxes this inductive bias (purple dotted arrows), giving the network greater capacity to attend across the full view during global attention.
}
\label{fig:dynamic_grouping}
\end{figure}

\begin{figure*}[t]
\centering
\includegraphics[width=1.0\textwidth]{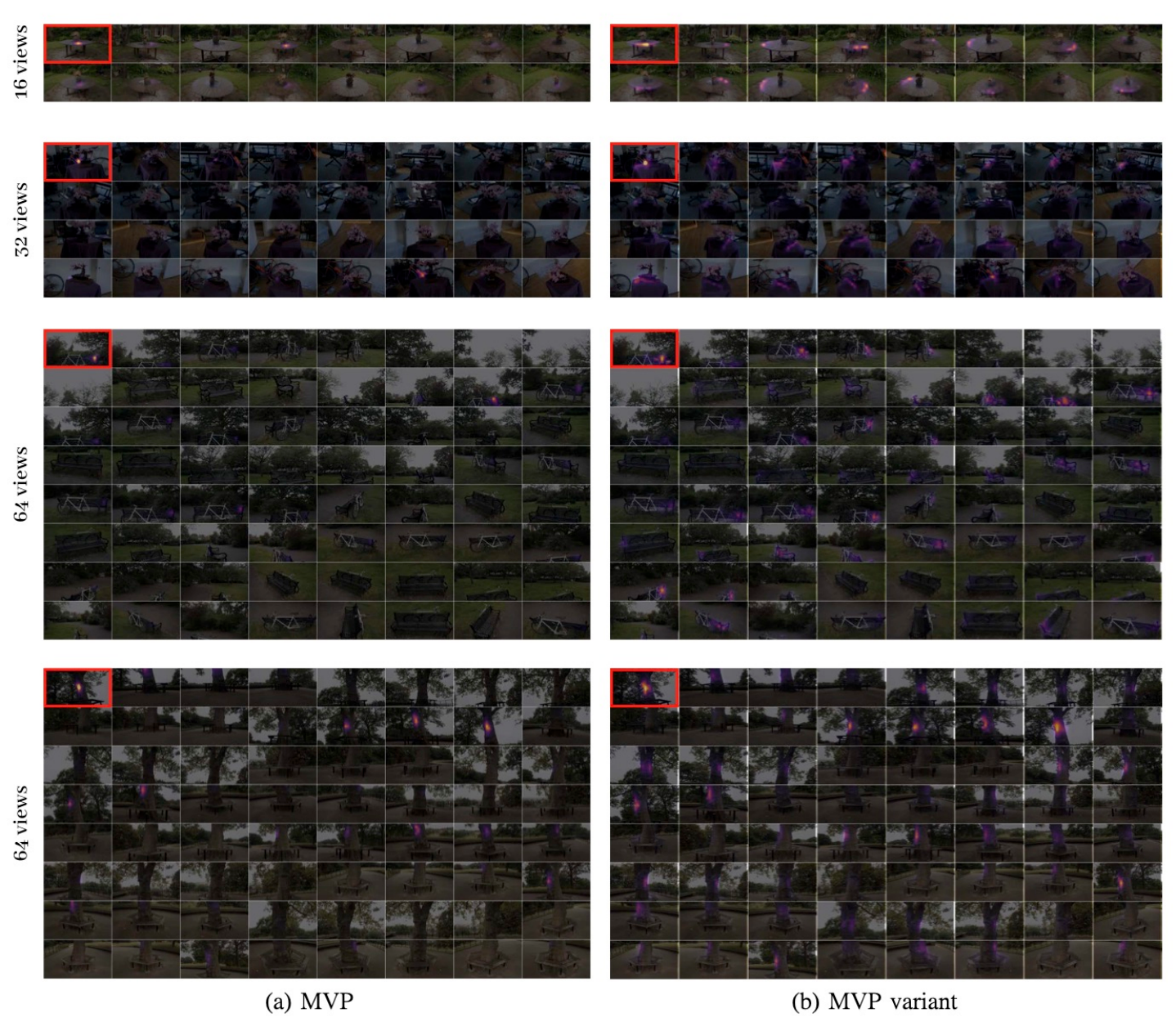}
\caption{Visualization of the averaged global attention map in the last layer for (a) MVP and (b) MVP variant. Given a patch in the reference frame (top-left, outlined in red), MVP variant attends to all relevant patches across every input frame, resulting in more 3D-consistent attention, whereas MVP attends to only a few.}
\label{fig:attn_heatmap}
\end{figure*}

\noindent\textbf{MVP Variants.} MVP performs well overall, but we find that its global attention lacks 3D-consistency, which can lead to low generalization performance (\cref{fig:attn_heatmap}(a)). We hypothesize that this stems from three limiting factors: fixed group size, network capacity, and dataset variety, as MVP is trained only on the DL3DV dataset. In particular, a fixed group size imposes a strong inductive bias that constrains the network's ability to reason globally, whereas dynamically varying group sizes can remove this bias (\cref{fig:dynamic_grouping}). Building on this insight, we introduce a more powerful MVP variant that combines our dynamic grouping mechanism with greater network capacity and a substantially larger training set of approximately 34M samples, while keeping training cost equivalent to that of the original model. Further details on this variant are provided in \cref{tab:mvpv2_info}. Thanks to this combination, it achieves highly 3D-consistent global attention (\cref{fig:attn_heatmap}(b)) and strong generalization performance (\cref{tab:mvpv2_generalization_nvs}), even surpassing the original MVP when integrated into our two-experts framework.

Although this MVP variant exhibits richer 3D consistency than MVP, we empirically find that this improved consistency does not always translate into stronger novel-view synthesis performance, as shown in Table~\ref{tab:mvpv2_generalization_nvs}. We attribute this to how the value of 3D consistency shifts with the amount of available input information. In the challenging sparse-view setting (e.g., 4, 8, or 16 views), 3D consistency plays a crucial role: with only a handful of observations, the model has little redundant information to draw on, so recovering accurate cross-view correspondences becomes essential for reconstructing correct geometry and, in turn, producing more accurate novel views. In the dense-view setting (e.g., 32, 64, or 128 views), however, this benefit diminishes. With many views available, overlapping information across the input images already provides strong cross-view constraints, so the model can compensate for weaker 3D consistency simply by aggregating redundant observations. As a result, this variant's more accurate 3D consistency no longer translates into a clear advantage, and MVP performs on par with, or even slightly better than, this variant in this regime.

\begin{table*}[t]
\centering
\small
\renewcommand{\arraystretch}{1.1}
\begin{tabular}{l|c|c}
\toprule
 & MVP & MVP variant \\
\midrule
Model size & 240M & 559M \\
\midrule
Training dataset & \makecell[c]{DL3DV~\cite{ling2024dl3dv}} & \makecell[c]{
DL3DV~\cite{ling2024dl3dv} \\
RE10K~\cite{zhou2018stereo} \\
WildRGBD~\cite{xia2024rgbd} \\
CO3Dv2~\cite{reizenstein21co3d} \\
MegaSynth~\cite{jiang2025megasynth} \\
Syn4D~\cite{jiang2026syn4d} \\
ScanNet++v2~\cite{yeshwanth2023scannet++} \\
BlendedMVS~\cite{yao2020blendedmvs} \\
SWORD~\cite{solovev2023self} \\
UnrealStereo4K~\cite{Tosi2021CVPR} \\
MVS-Synth~\cite{DeepMVS}
} \\
\midrule
Training resolutions & \makecell[c]{(480, 270) \\ (960, 540)} & \makecell[c]{
(512, 512) \\
(512, 384) \\
(512, 336) \\
(512, 288) \\
(384, 512) \\
(336, 512) \\
(288, 512) \\
(688, 512) \\
(768, 512) \\
(912, 512) \\
(512, 688) \\
(512, 768) \\
(512, 912)
} \\
\midrule
Grouping mechanism & Fixed grouping & Dynamic grouping \\
\bottomrule
\end{tabular}
\caption{Key differences between the original MVP and the its variant used as the appearance expert.}
\label{tab:mvpv2_info}
\end{table*}

\section{Limitations}
Our two-experts design introduces two limitations. First, because the appearance expert carries most of the rendering quality, overall performance is bottlenecked by its capacity, regardless of improvements elsewhere in the framework. Second, since pose supervision serves mainly as a regularization signal rather than a primary objective, our pose estimation accuracy is slightly lower than methods specifically designed for pose prediction.

\section{Additional Qualitative Results} 
\begin{figure*}[t]
\centering
\includegraphics[width=1.0\textwidth]{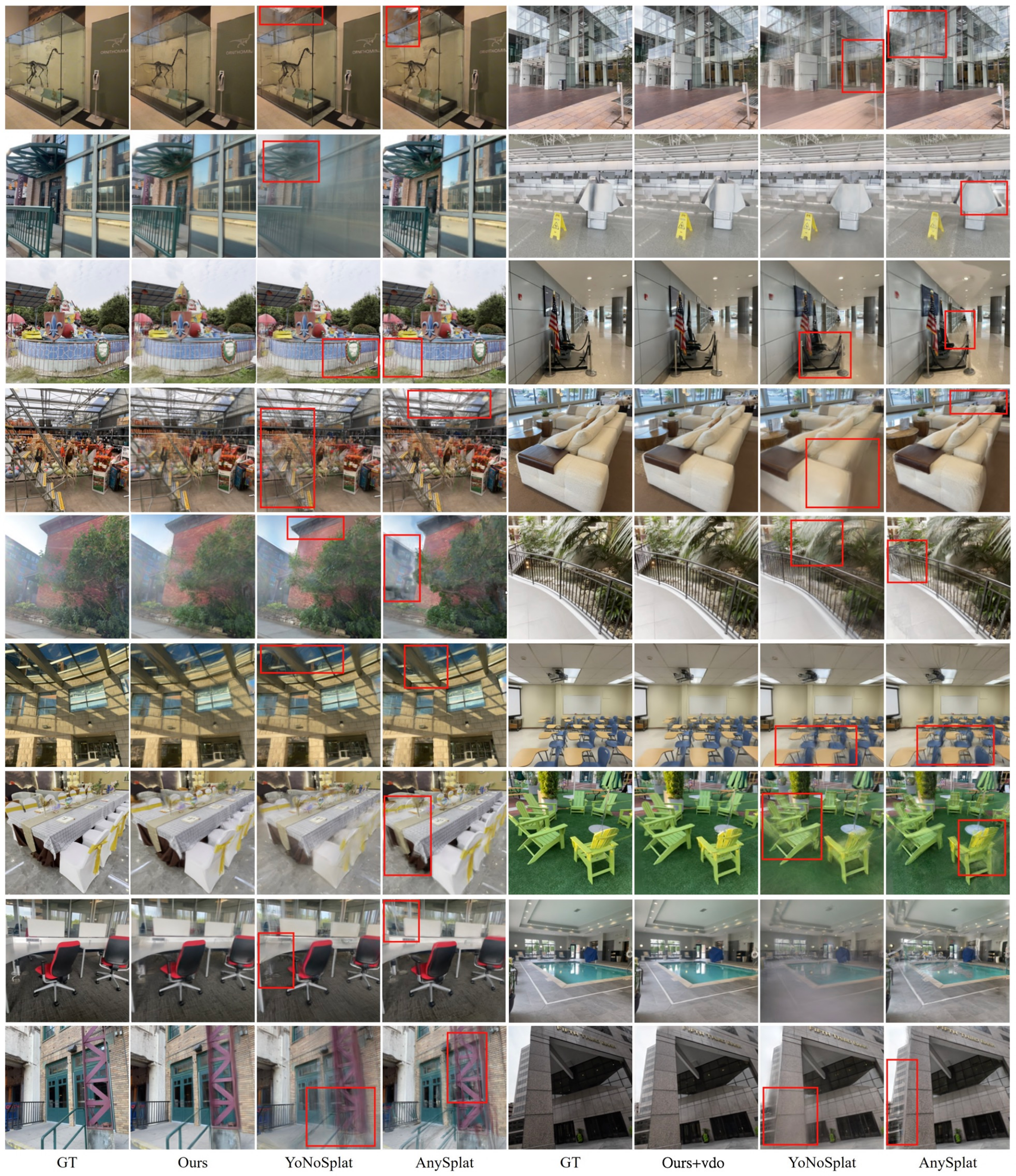}
\caption{Qualitative comparison on DL3DV ($224 \times 224$ resolution, 1-3 rows: 6 context views, 4-6 rows: 12 context views, 7-9 rows: 24 context views).}
\label{fig:supp_dl3dv}
\end{figure*}

\begin{figure*}[t]
\centering
\includegraphics[width=0.8\textwidth]{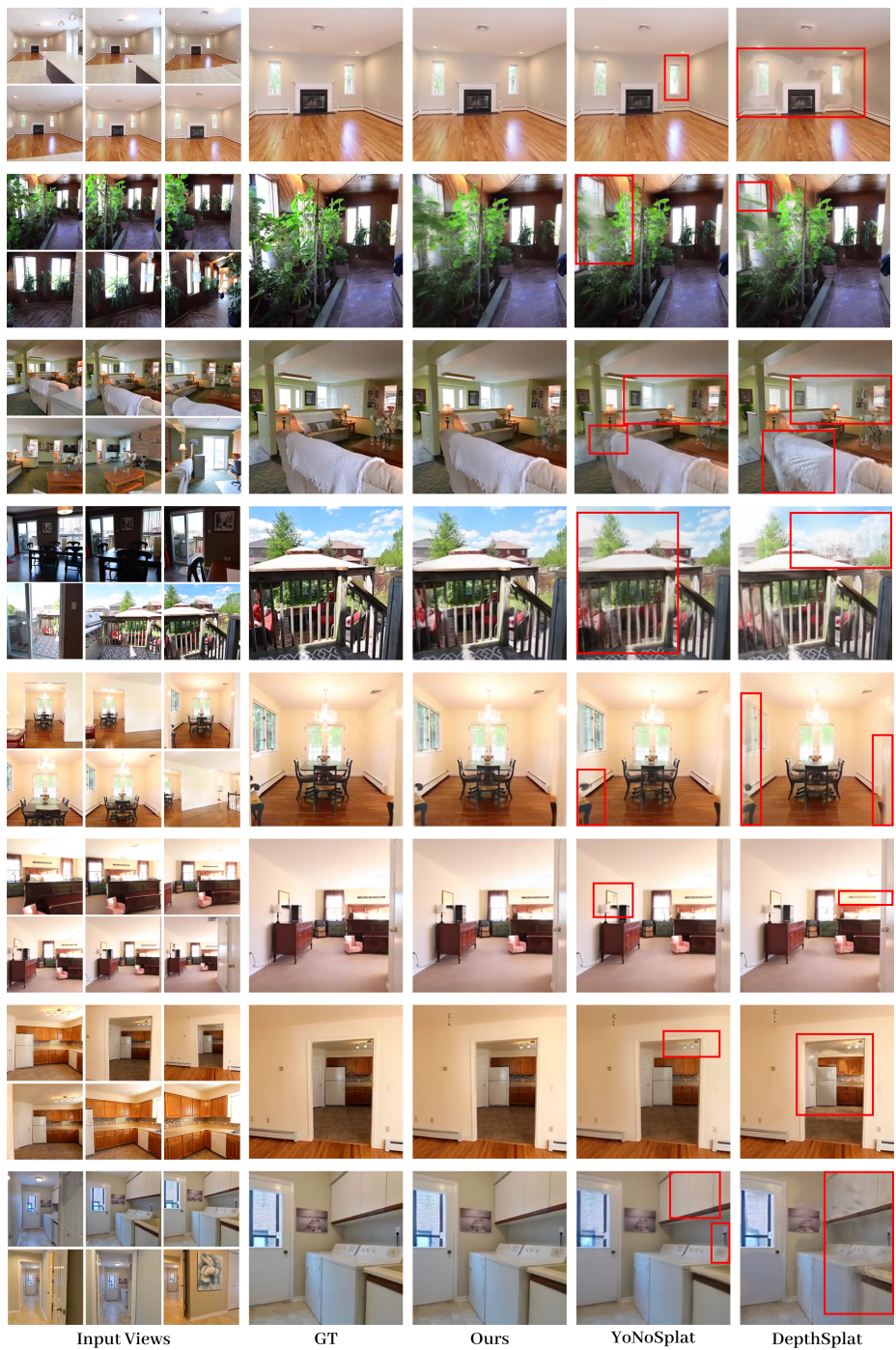}
\caption{Qualitative comparison on RE10K dataset with 6 context views ($224 \times 224$ resolutions).}
\label{fig:supp_re10k}
\end{figure*}

\begin{figure*}[t]
\centering
\includegraphics[width=1.0\textwidth]{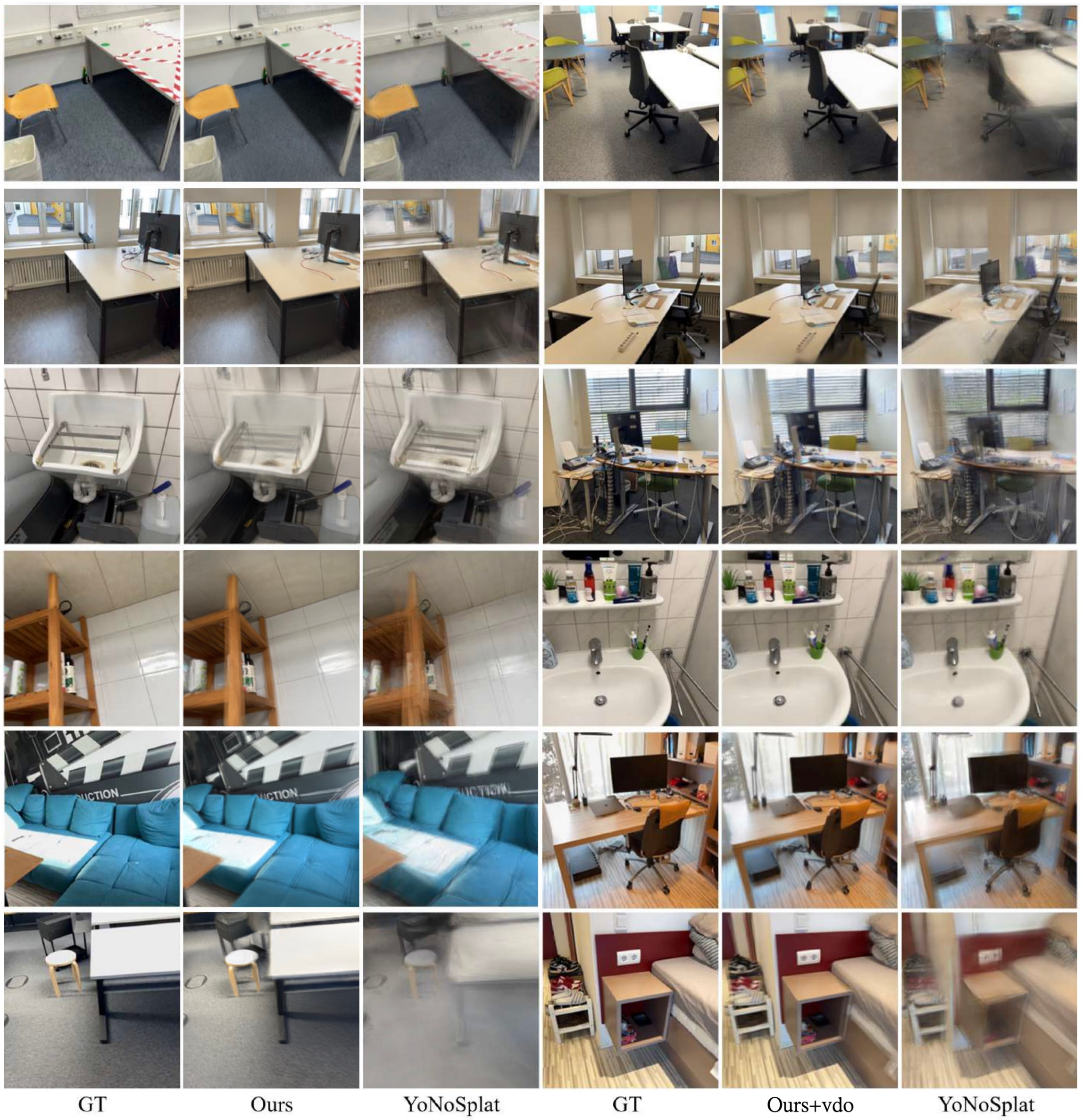}
\caption{Qualitative cross-dataset generalization results from DL3DV to ScanNet++ ($224 \times 224$ resolution, 1-2 rows: 32 context views, 3-4 rows: 64 context views, 5-6 rows: 128 context views).}
\label{fig:supp_scannet}
\end{figure*}

\begin{figure*}[t]
\centering
\includegraphics[width=1.0\textwidth]{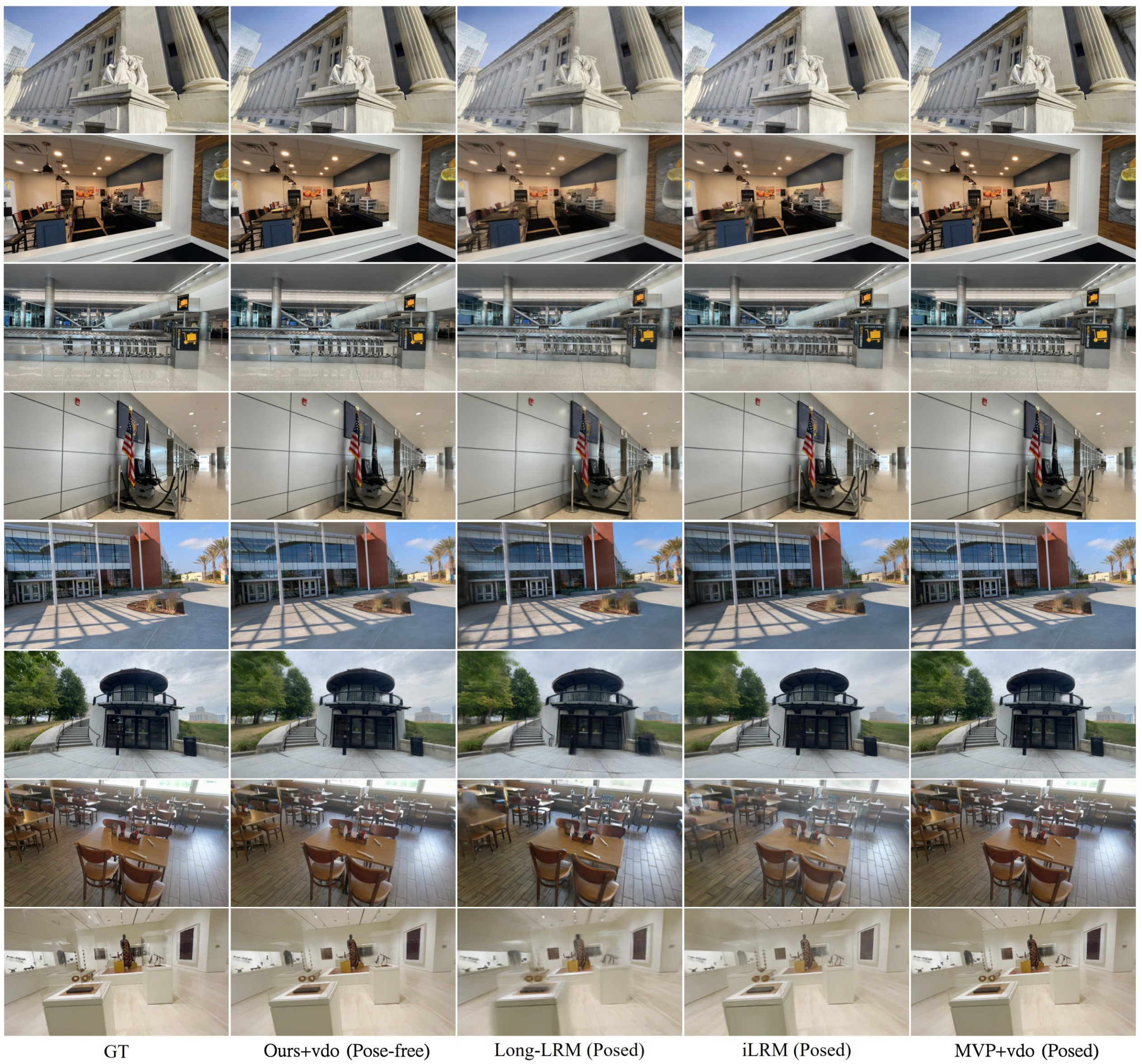}
\caption{Qualitative comparison on high-resolution DL3DV ($960 \times 540$, 1-2 rows: 16 context views, 3-4 rows: 32 context views, 5-6 rows: 64 context views, 7-8 rows: 128 context views).}
\label{fig:supp_dl3dv_hr}
\end{figure*}

We provide additional qualitative comparisons on RE10K~\cite{zhou2018stereo}, DL3DV~\cite{ling2024dl3dv}, high-resolution DL3DV, and ScanNet++~\cite{yeshwanth2023scannet++} (\cref{fig:supp_re10k,fig:supp_dl3dv,fig:supp_scannet,fig:supp_dl3dv_hr}), highlighting strong reconstruction quality across datasets and resolutions.

\begin{table*}[t]
\centering
\setlength{\tabcolsep}{4pt}
\renewcommand{\arraystretch}{1.2}
\begin{tabular}{lcccccccccc}
\toprule
Method & Views & \multicolumn{3}{c}{Tanks \& Temples} & \multicolumn{3}{c}{Mip-NeRF360} & \multicolumn{3}{c}{ETH3D} \\
\cmidrule(lr){3-5}\cmidrule(lr){6-8}\cmidrule(lr){9-11}
& & PSNR $\uparrow$ & SSIM $\uparrow$ & LPIPS $\downarrow$ & PSNR $\uparrow$ & SSIM $\uparrow$ & LPIPS $\downarrow$ & PSNR $\uparrow$ & SSIM $\uparrow$ & LPIPS $\downarrow$ \\
\midrule
Ours+vdo$^{\dagger}$ & 4 & 13.96 & 0.494 & 0.604 & 14.75 & 0.375 & 0.682 & 13.61 & 0.498 & 0.650 \\
\rowcolor{gray!20}
Ours+vdo$^{\dagger}$ &  & 14.33 & 0.491 & 0.600 & 16.86 & 0.389 & 0.628 & 15.81 & 0.522 & 0.623 \\
\midrule
Ours+vdo$^{\dagger}$ & 8 & 15.12 & 0.534 & 0.516 & 17.48 & 0.434 & 0.586 & 16.75 & 0.547 & 0.555 \\
\rowcolor{gray!20}
Ours+vdo$^{\dagger}$ &  & 15.50 & 0.530 & 0.509 & 18.34 & 0.432 & 0.538 & 17.44 & 0.552 & 0.532 \\
\midrule
Ours+vdo$^{\dagger}$ & 16 & 17.49 & 0.597 & 0.402 & 19.90 & 0.497 & 0.476 & 19.51 & 0.585 & 0.454 \\
\rowcolor{gray!20}
Ours+vdo$^{\dagger}$ &  & 17.43 & 0.577 & 0.408 & 19.86 & 0.475 & 0.457 & 19.21 & 0.572 & 0.456 \\
\midrule
Ours+vdo$^{\dagger}$ & 32 & 19.14 & 0.663 & 0.301 & 21.15 & 0.509 & 0.405 & 18.41 & 0.536 & 0.423 \\
\rowcolor{gray!20}
Ours+vdo$^{\dagger}$ &  & 18.80 & 0.635 & 0.324 & 21.02 & 0.508 & 0.401 & 18.22 & 0.521 & 0.456 \\
\midrule
Ours+vdo$^{\dagger}$ & 64 & 20.65 & 0.710 & 0.256 & 21.85 & 0.535 & 0.375 & 15.92 & 0.371 & 0.467 \\
\rowcolor{gray!20}
Ours+vdo$^{\dagger}$ &  & 19.36 & 0.637 & 0.313 & 21.91 & 0.542 & 0.368 & 16.18 & 0.350 & 0.477 \\
\midrule
Ours+vdo$^{\dagger}$ & 128 & 21.59 & 0.743 & 0.227 & 22.34 & 0.557 & 0.258 & - & - & - \\
\rowcolor{gray!20}
Ours+vdo$^{\dagger}$ &  & 20.11 & 0.668 & 0.287 & 22.33 & 0.559 & 0.340 & - & - & - \\
\bottomrule
\end{tabular}
\caption{\textbf{Cross Dataset Generalization (Tanks \& Temples, Mip-NeRF360, ETH3D).} The gray-colored row denotes \textit{2Xplat} results using MVP variant as the appearance backbone, while the non-gray row uses the original MVP. All context and target views are sampled uniformly from the original dataset resolution. Results for 128 views on ETH3D are omitted due to an insufficient number of available input views.}
\label{tab:cross_dataset_generalization}
\end{table*}

\clearpage
\bibliography{aaai2027}

\end{document}

%% file: Tables/dl3dv_nvs.tex
\begin{table*}[t]
\centering
\small
\setlength{\tabcolsep}{9pt}
\renewcommand{\arraystretch}{1.0}

\begin{tabular}{@{}lcc|ccc|ccc|ccc@{}}
\toprule
\multirow{2}{*}{Method}
& \multirow{2}{*}{$p$}
& \multirow{2}{*}{$k$}
& \multicolumn{3}{c|}{6v}
& \multicolumn{3}{c|}{12v}
& \multicolumn{3}{c}{24v} \\
\cmidrule(lr){4-6}
\cmidrule(lr){7-9}
\cmidrule(lr){10-12}

& &
& PSNR$\uparrow$ & SSIM$\uparrow$ & LPIPS$\downarrow$
& PSNR$\uparrow$ & SSIM$\uparrow$ & LPIPS$\downarrow$
& PSNR$\uparrow$ & SSIM$\uparrow$ & LPIPS$\downarrow$ \\
\midrule

MVSplat
& \checkmark & \checkmark
& 22.659 & 0.760 & 0.173
& 21.289 & 0.709 & 0.224
& 19.975 & 0.662 & 0.269 \\

DepthSplat
& \checkmark & \checkmark
& 23.418 & 0.797 & 0.136
& 21.911 & 0.753 & 0.179
& 20.088 & 0.690 & 0.240 \\

YoNoSplat
& \checkmark & \checkmark
& 24.717 & 0.817 & 0.139
& 23.285 & 0.773 & 0.177
& 22.664 & 0.758 & 0.192 \\

Ours & \checkmark & \checkmark
& \textbf{25.217} & \textbf{0.827} & \textbf{0.150}
& \textbf{24.592} & \textbf{0.813} & \textbf{0.169}
& \textbf{24.354} & \textbf{0.811} & \textbf{0.174} \\

Ours+vdo
& \checkmark & \checkmark
& 26.631 & 0.856 & 0.121 
& 27.240 & 0.866 & 0.115 
& 27.413 & 0.877 & 0.109 \\

\midrule

NoPoSplat$^{\dagger}$
& & \checkmark
& 22.766 & 0.743 & 0.179
& 19.380 & 0.563 & 0.318
& 17.860 & 0.495 & 0.397 \\

YoNoSplat$^{\dagger}$
& & \checkmark
& 24.887 & 0.819 & 0.138
& 23.149 & 0.758 & 0.183
& 22.354 & 0.731 & 0.205 \\

Ours$^{\dagger}$ &  & \checkmark
& \textbf{25.210} & \textbf{0.825} & \textbf{0.152}
& \textbf{24.337} & \textbf{0.798} & \textbf{0.177}
& \textbf{24.049} & \textbf{0.792} & \textbf{0.188} \\

Ours+vdo$^{\dagger}$
& & \checkmark
& 26.673 & 0.855 & 0.122
& 26.971 & 0.855 & 0.122
& 27.094 & 0.865 & 0.116 \\

\midrule

AnySplat
& &
& 19.027 & 0.554 & 0.235
& 18.940 & 0.549 & 0.262
& 19.703 & 0.596 & 0.249 \\

YoNoSplat
& &
& 22.290 & 0.695 & 0.173
& 20.383 & 0.602 & 0.229
& 19.711 & 0.572 & 0.255 \\

YoNoSplat$^{\dagger}$
& &
& 24.531 & 0.804 & 0.142
& 22.933 & 0.746 & 0.187
& 22.174 & 0.720 & 0.209 \\

Ours &  & 
& 24.835 & 0.814 & 0.154 
& 23.851 & 0.778 & 0.182
& 23.545 & 0.771 & 0.191 \\

Ours$^{\dagger}$ &  & 
& \textbf{25.227} & \textbf{0.825} & \textbf{0.151}
& \textbf{24.360} & \textbf{0.799} & \textbf{0.177}
& \textbf{24.077} & \textbf{0.793} & \textbf{0.185} \\

Ours+vdo
& &
& 26.007 & 0.839 & 0.126
& 26.015 & 0.826 & 0.129
& 25.894 & 0.832 & 0.125 \\

Ours+vdo$^{\dagger}$
& &
& 26.670 & 0.855 & 0.122
& 26.963 & 0.854 & 0.121
& 27.083 & 0.865 & 0.116 \\

\bottomrule
\end{tabular}
\normalsize
\caption{Novel view synthesis results on DL3DV with 6, 12, and 24 input views at low-resolution ($224 \times 224$). $p$ and $k$ denote the use of ground-truth poses and intrinsics, respectively. $^{\dagger}$ indicates evaluation with EPA. `vdo' denotes the use of view-dependent opacity and regularization in the appearance expert, and this variant is excluded from bold-letter ranking for fair comparison with baselines.}

\label{tab:nvs_dl3dv}
\end{table*}

%% file: Tables/dl3dv_hr.tex
\begin{table*}[t]
    \centering
    \setlength{\tabcolsep}{2pt}
    \renewcommand{\arraystretch}{1.0}
    \small
    \begin{tabular}{@{}lcccccccccccccc@{}}
    \toprule
    \multirow{2}{*}{Method} &\multirow{2}{*}{Optim.} &\multirow{2}{*}{(p, k)} & \multicolumn{3}{c}{16v} & \multicolumn{3}{c}{32v} & \multicolumn{3}{c}{64v} & \multicolumn{3}{c}{128v} \\
    \cmidrule(lr){4-6}\cmidrule(lr){7-9}\cmidrule(lr){10-12}\cmidrule(lr){13-15}
    & & & PSNR $\uparrow$ & SSIM $\uparrow$ & LPIPS $\downarrow$
    & PSNR $\uparrow$ & SSIM $\uparrow$ & LPIPS $\downarrow$
    & PSNR $\uparrow$ & SSIM $\uparrow$ & LPIPS $\downarrow$
    & PSNR $\uparrow$ & SSIM $\uparrow$ & LPIPS $\downarrow$ \\
    \midrule
    3D-GS$_{30k}$ 
    & \checkmark & \checkmark & 21.48 & 0.753 & 0.252 
    & 24.43 & 0.827 & 0.191 
    & 27.34 & 0.883 & 0.146 
    & 29.43 & 0.914 & 0.123 \\
    \midrule
    Long-LRM
    & & \checkmark & 21.05 & 0.708 & 0.297 
    & 23.97 & 0.778 & 0.267 
    & 23.60 & 0.789 & 0.260 
    & 21.24 & 0.739 & 0.308 \\
    iLRM
    & & \checkmark & 21.92 & 0.748 & 0.316 
    & 24.30 & 0.803 & 0.256 
    & 24.44 & 0.819 & 0.240 
    & 22.98 & 0.807 & 0.249 \\
    MVP+vdo
    & & \checkmark & 23.76 & 0.798 & 0.239
    & 25.96 & 0.847 & 0.187
    & 27.73 & 0.881 & 0.154 
    & 29.02 & 0.903 & 0.134 \\
    \midrule
    Ours+vdo
    & & & {22.90} & {0.754} & {0.259} 
    & {24.75} & {0.800} & {0.208} 
    & {26.11} & {0.830} & {0.180} 
    & {27.16} & {0.853} & {0.162} \\       
    Ours+vdo$^{\dagger}$
    & & & {23.61} & {0.786} & {0.248} 
    & {25.66} & {0.832} & {0.198} 
    & {27.12} & {0.859} & {0.171} 
    & {28.30} & {0.880} & {0.153} \\   
    \bottomrule
    \end{tabular}
    \normalsize
    \caption{Quantitative comparison on the DL3DV dataset under varying numbers of input views (16, 32, 64, and 128) for high-resolution ($960 \times 540$) novel view synthesis.}
    \label{tab:quantitative_result_on_dl3dv_hr}
\end{table*}

%% file: Tables/re10k_nvs.tex
\begin{table}[t]
\centering
\small
\setlength{\tabcolsep}{3pt}
\renewcommand{\arraystretch}{0.8}
\begin{tabular}{lcc|ccc}
\toprule
Method & $p$ & $k$ & PSNR$\uparrow$ & SSIM$\uparrow$ & LPIPS$\downarrow$ \\
\midrule
DepthSplat
& \checkmark & \checkmark
& 24.156 & 0.846 & 0.145 \\
YoNoSplat
& \checkmark & \checkmark
& 25.037 & 0.848 & 0.134 \\
\midrule
NoPoSplat$^{\dagger}$
& & \checkmark
& 22.175 & 0.750 & 0.207 \\
YoNoSplat$^{\dagger}$
& & \checkmark
& 25.395 & 0.857 & 0.131 \\
Ours$^{\dagger}$
& & \checkmark
& \textbf{27.108} & \textbf{0.877} & \textbf{0.128} \\
\midrule
YoNoSplat
& & 
& 19.723 & 0.613 & 0.229 \\
YoNoSplat$^{\dagger}$
& &
& 24.571 & 0.823 & 0.144 \\
Ours
& &
& 26.161 & 0.859 & 0.132 \\
Ours$^{\dagger}$
& &
& \textbf{27.239} & \textbf{0.881} & \textbf{0.126} \\
\bottomrule
\end{tabular}
\caption{Quantitative comparison on low-resolution RE10K dataset with 6 context views.}
\label{tab:re10k_nvs}
\end{table}

%% file: Tables/dl3dv_2_scannetpp_nvs.tex
\begin{table}[t]
\centering
\footnotesize
\setlength{\tabcolsep}{4pt}
\begin{tabular}{l|c|c|c}
\toprule
Method & 32v & 64v & 128v \\
\midrule
AnySplat         & 14.05 & 15.98 & 16.99 \\
YoNoSplat$^{\dagger}$ w/o GT $k$  & 16.89 & 17.37 & 17.64 \\
YoNoSplat$^{\dagger}$ w/ GT $k$   & 17.94 & 18.83 & 19.28 \\
Ours$^{\dagger}$ w/o GT $k$ & \textbf{18.04} & \textbf{19.64} & \textbf{20.74} \\
Ours$^{\dagger}$ w/ GT $k$ & \textbf{18.09} & \textbf{19.74} & \textbf{20.88} \\
\bottomrule
\end{tabular}
\caption{Cross-dataset generalization from DL3DV to ScanNet++ (PSNR$\uparrow$).}
\label{tab:dl3dv_2_scannetpp}
\end{table}

%% file: Tables/dl3dv_sparse_view.tex
\begin{table}[t]
\centering
\footnotesize
\setlength{\tabcolsep}{4pt}
\begin{tabular}{l|ccc}
\toprule
Method & PSNR$\uparrow$ & SSIM$\uparrow$ & LPIPS$\downarrow$ \\
\midrule
FLARE & 20.07 & 0.606 & 0.315 \\
NoPoSplat$^{\dagger}$ & 20.36 & 0.636 & 0.291 \\
YoNoSplat & 18.68 & 0.526 & 0.337 \\
YoNoSplat$^{\dagger}$ & 20.22 & 0.617 & 0.304 \\
Ours & 19.56 & 0.574 & 0.337 \\
Ours$^{\dagger}$ & 20.06 & 0.606 & 0.329 \\ 
\bottomrule
\end{tabular}
\normalsize
\caption{Sparse-view (two context views) NVS on DL3DV.}
\label{tab:sparse_view_ablation}
\end{table}

\begin{table}[t]
\centering
\footnotesize
\setlength{\tabcolsep}{3pt}
\begin{tabular}{l|c|c|cc}
\toprule
Method & size & speed & PSNR$\uparrow$ & LPIPS$\downarrow$ \\
\midrule
DA3-G$^{\dagger}$ & 1.1B & 0.65s & 23.77 & 0.165 \\
YoNoSplat$^{\dagger}$ ($\pi^{3}$) & 1B & 0.33s & 24.53 & 0.142 \\
\midrule
Ours$^{\dagger}$ (DA3-L) & 0.5B & 0.15s & 24.93 & 0.160 \\
Ours$^{\dagger}$ (DA3-G) & 1.3B & 0.31s & 25.23 & 0.151 \\
Ours$^{\dagger}$ ($\pi^{3}$) & 1.1B & 0.27s & 25.08 & 0.156 \\
\bottomrule
\end{tabular}
\normalsize
\caption{Efficiency trade-off (RTX 3090 inference speed).}
\label{tab:size_comparison}
\end{table}

%% file: Tables/monolithic_vs_twoexperts.tex
\begin{table}[!h]
\centering
\footnotesize
\setlength{\tabcolsep}{4pt}
\begin{tabular}{l|l|c|c|c}
\toprule
Backbone & Method & 6v & 12v & 24v \\
\midrule
\multirow{2}{*}{DA3-G}
& DA3-G$^{\dagger}$ 
& 23.77 & 22.38 & 21.69 \\
& Ours$^{\dagger}$   
& \textbf{25.23} & \textbf{24.36} & \textbf{24.08} \\
\midrule
\multirow{2}{*}{DA3-L}
& DA3-L$^{\dagger}$ 
& 21.87 & 20.93 & 20.61 \\
& Ours$^{\dagger}$
& \textbf{24.93} & \textbf{23.87} & \textbf{23.48} \\
\midrule
\multirow{4}{*}{$\pi^3$}
& YoNoSplat            
& 22.29 & 20.38 & 19.71 \\
& YoNoSplat$^{\dagger}$ 
& 24.53 & 22.93 & 22.17 \\
& Ours  
& 24.52 & 23.76 & 23.12 \\
& Ours$^{\dagger}$   
& \textbf{25.08} & \textbf{24.32} & \textbf{23.83} \\
\bottomrule
\end{tabular}
\normalsize
\caption{Monolithic vs. two-experts architectures (PSNR$\uparrow$).}
\label{tab:monolithic_vs_twoexperts}
\end{table}

%% file: Tables/joint_training_ablation.tex
\begin{table}[t]
\centering
\footnotesize
\setlength{\tabcolsep}{4pt}
\begin{tabular}{c|cc|cc|cc}
\toprule
\multirow{2}{*}{}
& \multicolumn{2}{c|}{6v}
& \multicolumn{2}{c|}{12v}
& \multicolumn{2}{c}{24v} \\
\cmidrule(lr){2-3} \cmidrule(lr){4-5} \cmidrule(lr){6-7}
& PSNR$\uparrow$ & LPIPS$\downarrow$
& PSNR$\uparrow$ & LPIPS$\downarrow$
& PSNR$\uparrow$ & LPIPS$\downarrow$ \\
\midrule
$\times$$^{\dagger}$ 
& 21.04 & 0.267
& 19.96 & 0.322
& 19.38 & 0.360 \\
$\circ$$^{\dagger}$ 
& 25.23 & 0.151
& 24.36 & 0.177
& 24.08 & 0.185 \\
\bottomrule
\end{tabular}
\caption{Joint training analysis on DL3DV $224 \times 224$.}
\label{tab:joint_training_ablation}
\end{table}

%% file: Tables/appearance_expert_ablation.tex
\begin{table}[!h]
\centering
\footnotesize
\setlength{\tabcolsep}{6pt}
\begin{tabular}{l|c|ccc}
\toprule
Method & (p, k) & 16v & 32v & 64v \\
\midrule
iLRM & \checkmark & 21.92 & 24.30 & 24.44 \\
Ours$^\dagger$ (iLRM) & & 19.71 & 22.84 & 22.10 \\
Long-LRM & \checkmark & 21.05 & 23.97 & 23.60 \\
Ours$^\dagger$ (Long-LRM) & & 20.26 & 22.41 & 22.06 \\
\bottomrule
\end{tabular}
\caption{Generality across appearance experts (PSNR$\uparrow$).}
\label{tab:appearance_expert_ablation}
\end{table}

%% file: Tables/mvpv2_generalization_nvs.tex
\begin{table*}[t]
\centering
\setlength{\tabcolsep}{2pt}
\renewcommand{\arraystretch}{1.0}
\begin{tabular}{@{}lccccccccccccc@{}}
\toprule
\multicolumn{13}{l}{\textbf{(a) 4, 8, 16 input views}} \\
\midrule
\multirow{2}{*}{Method} & \multicolumn{3}{c}{Tanks \& Temples} & \multicolumn{3}{c}{Mip-NeRF360} & \multicolumn{3}{c}{ETH3D} & \multicolumn{3}{c}{Average}\\
\cmidrule(lr){2-4}\cmidrule(lr){5-7}\cmidrule(lr){8-10}\cmidrule(lr){11-13}
& PSNR $\uparrow$ & SSIM $\uparrow$ & LPIPS $\downarrow$
& PSNR $\uparrow$ & SSIM $\uparrow$ & LPIPS $\downarrow$
& PSNR $\uparrow$ & SSIM $\uparrow$ & LPIPS $\downarrow$
& PSNR $\uparrow$ & SSIM $\uparrow$ & LPIPS $\downarrow$ \\
\midrule
Ours+vdo$^{\dagger}$
& 15.52 & 0.542 & 0.507 
& 17.38 & 0.435 & 0.581 
& 16.62 & 0.543 & 0.553
& 16.51 & \textbf{0.507} & 0.547 \\   
\rowcolor{gray!20}      
Ours+vdo$^{\dagger}$
& 15.75 & 0.533 & 0.506
& 18.35 & 0.432 & 0.541
& 17.49 & 0.549 & 0.537
& \textbf{17.20} & 0.505 & \textbf{0.528} \\   
\midrule
\midrule
\multicolumn{13}{l}{\textbf{(b) 32, 64, 128 input views}} \\
\midrule
\multirow{2}{*}{Method} & \multicolumn{3}{c}{Tanks \& Temples} & \multicolumn{3}{c}{Mip-NeRF360} & \multicolumn{3}{c}{ETH3D} & \multicolumn{3}{c}{Average}\\
\cmidrule(lr){2-4}\cmidrule(lr){5-7}\cmidrule(lr){8-10}\cmidrule(lr){11-13}
& PSNR $\uparrow$ & SSIM $\uparrow$ & LPIPS $\downarrow$
& PSNR $\uparrow$ & SSIM $\uparrow$ & LPIPS $\downarrow$
& PSNR $\uparrow$ & SSIM $\uparrow$ & LPIPS $\downarrow$
& PSNR $\uparrow$ & SSIM $\uparrow$ & LPIPS $\downarrow$ \\
\midrule
Ours+vdo$^{\dagger}$
& 20.46 & 0.705 & 0.261 
& 21.78 & 0.534 & 0.346 
& 17.17 & 0.454 & 0.445
& \textbf{19.80} & \textbf{0.564} & \textbf{0.351} \\   
\rowcolor{gray!20}      
Ours+vdo$^{\dagger}$
& 19.42 & 0.647 & 0.308
& 21.75 & 0.536 & 0.370
& 17.20 & 0.436 & 0.467
& 19.46 & 0.540 & 0.382 \\ 
\bottomrule
\end{tabular}
\caption{Average NVS generalization results across two input-view-count regimes: (a) 4, 8, and 16 input views, and (b) 32, 64, and 128 input views, evaluated on three datasets in the high-resolution setting ($960 \times 540$). The last three columns report the mean PSNR, SSIM, and LPIPS across the three datasets. The gray-colored row denotes \textit{2Xplat} results using MVP variant as the appearance backbone, while the non-gray row uses the original MVP. Full results are provided in \cref{tab:cross_dataset_generalization}.}
\label{tab:mvpv2_generalization_nvs}
\end{table*}

%% file: aaai2027.bib
@String(CVPR  = {IEEE Conf. Comput. Vis. Pattern Recog.})

@String(ICCV  = {Int. Conf. Comput. Vis.})

@String(ECCV  = {Eur. Conf. Comput. Vis.})

@String(TOG   = {ACM Trans. Graph.})

@String(CVPR  = {CVPR})

@String(ICCV  = {ICCV})

@String(ECCV  = {ECCV})

@String(TOG   = {ACM TOG})

@inproceedings{wang2026pi,
  title={$\pi3$: Permutation-Equivariant Visual Geometry Learning},
  author={Wang, Yifan and Zhou, Jianjun and Zhu, Haoyi and Chang, Wenzheng and Zhou, Yang and Li, Zizun and Chen, Junyi and Pang, Jiangmiao and Shen, Chunhua and He, Tong},
  booktitle={International Conference on Learning Representations},
  volume={2026},
  pages={10481--10497},
  year={2026}
}

@article{kerbl20233d,
  title={3d gaussian splatting for real-time radiance field rendering.},
  author={Kerbl, Bernhard and Kopanas, Georgios and Leimk{\"u}hler, Thomas and Drettakis, George and others},
  journal={ACM Trans. Graph.},
  volume={42},
  number={4},
  pages={139--1},
  year={2023}
}

@inproceedings{jiang2024vrgs,
  title={Vr-gs: A physical dynamics-aware interactive gaussian splatting system in virtual reality},
  author={Jiang, Ying and Yu, Chang and Xie, Tianyi and Li, Xuan and Feng, Yutao and Wang, Huamin and Li, Minchen and Lau, Henry and Gao, Feng and Yang, Yin and others},
  booktitle={ACM SIGGRAPH 2024 conference papers},
  pages={1--1},
  year={2024}
}

@article{jiang2024dualgs,
  title={Robust dual gaussian splatting for immersive human-centric volumetric videos},
  author={Jiang, Yuheng and Shen, Zhehao and Hong, Yu and Guo, Chengcheng and Wu, Yize and Zhang, Yingliang and Yu, Jingyi and Xu, Lan},
  journal={ACM Transactions on Graphics (TOG)},
  volume={43},
  number={6},
  pages={1--15},
  year={2024},
  publisher={ACM New York, NY, USA}
}

@article{sun2024splatter,
  title={Splatter a video: Video gaussian representation for versatile processing},
  author={Sun, Yang-Tian and Huang, Yihua and Ma, Lin and Lyu, Xiaoyang and Cao, Yan-Pei and Qi, Xiaojuan},
  journal={Advances in Neural Information Processing Systems},
  volume={37},
  pages={50401--50425},
  year={2024}
}

@article{shen2025nutshell,
  title={Seeing world dynamics in a nutshell},
  author={Shen, Qiuhong and Yi, Xuanyu and Lin, Mingbao and Zhang, Hanwang and Yan, Shuicheng and Wang, Xinchao},
  journal={arXiv preprint arXiv:2502.03465},
  year={2025}
}

@inproceedings{zhou2024drivinggaussian,
  title={Drivinggaussian: Composite gaussian splatting for surrounding dynamic autonomous driving scenes},
  author={Zhou, Xiaoyu and Lin, Zhiwei and Shan, Xiaojun and Wang, Yongtao and Sun, Deqing and Yang, Ming-Hsuan},
  booktitle={Proceedings of the IEEE/CVF conference on computer vision and pattern recognition},
  pages={21634--21643},
  year={2024}
}

@inproceedings{hess2025splatad,
  title={Splatad: Real-time lidar and camera rendering with 3d gaussian splatting for autonomous driving},
  author={Hess, Georg and Lindstr{\"o}m, Carl and Fatemi, Maryam and Petersson, Christoffer and Svensson, Lennart},
  booktitle={Proceedings of the Computer Vision and Pattern Recognition Conference},
  pages={11982--11992},
  year={2025}
}

@inproceedings{lu2024manigaussian,
  title={Manigaussian: Dynamic gaussian splatting for multi-task robotic manipulation},
  author={Lu, Guanxing and Zhang, Shiyi and Wang, Ziwei and Liu, Changliu and Lu, Jiwen and Tang, Yansong},
  booktitle={European Conference on Computer Vision},
  pages={349--366},
  year={2024},
  organization={Springer}
}

@inproceedings{charatan2024pixelsplat,
  title={pixelsplat: 3d gaussian splats from image pairs for scalable generalizable 3d reconstruction},
  author={Charatan, David and Li, Sizhe Lester and Tagliasacchi, Andrea and Sitzmann, Vincent},
  booktitle={Proceedings of the IEEE/CVF conference on computer vision and pattern recognition},
  pages={19457--19467},
  year={2024}
}

@inproceedings{chen2024mvsplat,
  title={Mvsplat: Efficient 3d gaussian splatting from sparse multi-view images},
  author={Chen, Yuedong and Xu, Haofei and Zheng, Chuanxia and Zhuang, Bohan and Pollefeys, Marc and Geiger, Andreas and Cham, Tat-Jen and Cai, Jianfei},
  booktitle={European conference on computer vision},
  pages={370--386},
  year={2024},
  organization={Springer}
}

@inproceedings{xu2025depthsplat,
  title={Depthsplat: Connecting gaussian splatting and depth},
  author={Xu, Haofei and Peng, Songyou and Wang, Fangjinhua and Blum, Hermann and Barath, Daniel and Geiger, Andreas and Pollefeys, Marc},
  booktitle={Proceedings of the Computer Vision and Pattern Recognition Conference},
  pages={16453--16463},
  year={2025}
}

@inproceedings{schoenberger2016sfm,
    author={Sch\"{o}nberger, Johannes Lutz and Frahm, Jan-Michael},
    title={Structure-from-Motion Revisited},
    booktitle={Conference on Computer Vision and Pattern Recognition (CVPR)},
    year={2016},
}

@inproceedings{schoenberger2016mvs,
    author={Sch\"{o}nberger, Johannes Lutz and Zheng, Enliang and Pollefeys, Marc and Frahm, Jan-Michael},
    title={Pixelwise View Selection for Unstructured Multi-View Stereo},
    booktitle={European Conference on Computer Vision (ECCV)},
    year={2016},
}

@inproceedings{zhang2024gslrm,
  title={Gs-lrm: Large reconstruction model for 3d gaussian splatting},
  author={Zhang, Kai and Bi, Sai and Tan, Hao and Xiangli, Yuanbo and Zhao, Nanxuan and Sunkavalli, Kalyan and Xu, Zexiang},
  booktitle={European Conference on Computer Vision},
  pages={1--19},
  year={2024},
  organization={Springer}
}

@inproceedings{yu2024mip,
  title={Mip-splatting: Alias-free 3d gaussian splatting},
  author={Yu, Zehao and Chen, Anpei and Huang, Binbin and Sattler, Torsten and Geiger, Andreas},
  booktitle={Proceedings of the IEEE/CVF conference on computer vision and pattern recognition},
  pages={19447--19456},
  year={2024}
}

@article{escontrela2025gaussgym,
  title={Gaussgym: An open-source real-to-sim framework for learning locomotion from pixels},
  author={Escontrela, Alejandro and Kerr, Justin and Allshire, Arthur and Frey, Jonas and Duan, Rocky and Sferrazza, Carmelo and Abbeel, Pieter},
  journal={arXiv preprint arXiv:2510.15352},
  year={2025}
}

@inproceedings{ye2026yonosplat,
  title={Yonosplat: You only need one model for feedforward 3d gaussian splatting},
  author={Ye, Botao and Chen, Boqi and Xu, Haofei and Barath, Daniel and Pollefeys, Marc},
  booktitle={International Conference on Learning Representations},
  volume={2026},
  pages={39852--39871},
  year={2026}
}

@inproceedings{ye2025no,
  title={No pose, no problem: Surprisingly simple 3d gaussian splats from sparse unposed images},
  author={Ye, Botao and Liu, Sifei and Xu, Haofei and Li, Xueting and Pollefeys, Marc and Yang, Ming-Hsuan and Peng, Songyou},
  booktitle={International Conference on Learning Representations},
  volume={2025},
  pages={54009--54033},
  year={2025}
}

@inproceedings{wang2025vggt,
  title={Vggt: Visual geometry grounded transformer},
  author={Wang, Jianyuan and Chen, Minghao and Karaev, Nikita and Vedaldi, Andrea and Rupprecht, Christian and Novotny, David},
  booktitle={Proceedings of the Computer Vision and Pattern Recognition Conference},
  pages={5294--5306},
  year={2025}
}

@inproceedings{kang2025selfsplat,
  title={Selfsplat: Pose-free and 3d prior-free generalizable 3d gaussian splatting},
  author={Kang, Gyeongjin and Yoo, Jisang and Park, Jihyeon and Nam, Seungtae and Im, Hyeonsoo and Shin, Sangheon and Kim, Sangpil and Park, Eunbyung},
  booktitle={Proceedings of the Computer Vision and Pattern Recognition Conference},
  pages={22012--22022},
  year={2025}
}

@article{jiang2025anysplat,
  title={Anysplat: Feed-forward 3d gaussian splatting from unconstrained views},
  author={Jiang, Lihan and Mao, Yucheng and Xu, Linning and Lu, Tao and Ren, Kerui and Jin, Yichen and Xu, Xudong and Yu, Mulin and Pang, Jiangmiao and Zhao, Feng and others},
  journal={ACM Transactions on Graphics (TOG)},
  volume={44},
  number={6},
  pages={1--16},
  year={2025},
  publisher={ACM New York, NY, USA}
}

@inproceedings{he2020epipolar,
  title={Epipolar transformers},
  author={He, Yihui and Yan, Rui and Fragkiadaki, Katerina and Yu, Shoou-I},
  booktitle={Proceedings of the ieee/cvf conference on computer vision and pattern recognition},
  pages={7779--7788},
  year={2020}
}

@article{li2025prope,
    title={Cameras as Relative Positional Encoding},
    author={Li, Ruilong and Yi, Brent and Liu, Junchen and Gao, Hang and Ma, Yi and Kanazawa, Angjoo},
    journal={Advances in Neural Information Processing Systems},
    year={2025}
}

@inproceedings{xiong2023cape,
  title={Cape: Camera view position embedding for multi-view 3d object detection},
  author={Xiong, Kaixin and Gong, Shi and Ye, Xiaoqing and Tan, Xiao and Wan, Ji and Ding, Errui and Wang, Jingdong and Bai, Xiang},
  booktitle={Proceedings of the IEEE/CVF conference on computer vision and pattern recognition},
  pages={21570--21579},
  year={2023}
}

@inproceedings{miyato2024gta,
  title={Gta: A geometry-aware attention mechanism for multi-view transformers},
  author={Miyato, Takeru and Jaeger, Bernhard and Welling, Max and Geiger, Andreas},
  booktitle={International Conference on Learning Representations},
  volume={2024},
  pages={8172--8208},
  year={2024}
}

@article{wu2026rayrope,
  title={RayRoPE: Projective Ray Positional Encoding for Multi-view Attention},
  author={Wu, Yu and Jeon, Minsik and Chang, Jen-Hao Rick and Tuzel, Oncel and Tulsiani, Shubham},
  journal={arXiv preprint arXiv:2601.15275},
  year={2026}
}

@inproceedings{jiang2025rayzer,
  title={Rayzer: A self-supervised large view synthesis model},
  author={Jiang, Hanwen and Tan, Hao and Wang, Peng and Jin, Haian and Zhao, Yue and Bi, Sai and Zhang, Kai and Luan, Fujun and Sunkavalli, Kalyan and Huang, Qixing and others},
  booktitle={2025 IEEE/CVF International Conference on Computer Vision (ICCV)},
  pages={4918--4929},
  year={2025},
  organization={IEEE}
}

@inproceedings{lai2021videoae,
  title={Video autoencoder: self-supervised disentanglement of static 3d structure and motion},
  author={Lai, Zihang and Liu, Sifei and Efros, Alexei A and Wang, Xiaolong},
  booktitle={Proceedings of the IEEE/CVF International Conference on Computer Vision},
  pages={9730--9740},
  year={2021}
}

@inproceedings{wang2024dust3r,
  title={Dust3r: Geometric 3d vision made easy},
  author={Wang, Shuzhe and Leroy, Vincent and Cabon, Yohann and Chidlovskii, Boris and Revaud, Jerome},
  booktitle={Proceedings of the IEEE/CVF conference on computer vision and pattern recognition},
  pages={20697--20709},
  year={2024}
}

@inproceedings{leroy2024mast3r,
  title={Grounding image matching in 3d with mast3r},
  author={Leroy, Vincent and Cabon, Yohann and Revaud, J{\'e}r{\^o}me},
  booktitle={European conference on computer vision},
  pages={71--91},
  year={2024},
  organization={Springer}
}

@article{dosovitskiy2020vit,
  title={An image is worth 16x16 words: Transformers for image recognition at scale},
  author={Dosovitskiy, Alexey and Beyer, Lucas and Kolesnikov, Alexander and Weissenborn, Dirk and Zhai, Xiaohua and Unterthiner, Thomas and Dehghani, Mostafa and Minderer, Matthias and Heigold, Georg and Gelly, Sylvain and others},
  journal={arXiv preprint arXiv:2010.11929},
  year={2020}
}

@inproceedings{edstedt2024roma,
  title={Roma: Robust dense feature matching},
  author={Edstedt, Johan and Sun, Qiyu and B{\"o}kman, Georg and Wadenb{\"a}ck, M{\aa}rten and Felsberg, Michael},
  booktitle={Proceedings of the IEEE/CVF conference on computer vision and pattern recognition},
  pages={19790--19800},
  year={2024}
}

@article{lin2025depth3,
  title={Depth anything 3: Recovering the visual space from any views},
  author={Lin, Haotong and Chen, Sili and Liew, Junhao and Chen, Donny Y and Li, Zhenyu and Shi, Guang and Feng, Jiashi and Kang, Bingyi},
  journal={arXiv preprint arXiv:2511.10647},
  year={2025}
}

@inproceedings{hong2024lrm,
  title={Lrm: Large reconstruction model for single image to 3d},
  author={Hong, Yicong and Zhang, Kai and Gu, Jiuxiang and Bi, Sai and Zhou, Yang and Liu, Difan and Liu, Feng and Sunkavalli, Kalyan and Bui, Trung and Tan, Hao},
  booktitle={International Conference on Learning Representations},
  volume={2024},
  pages={50678--50702},
  year={2024}
}

@inproceedings{fridovich2023kplane,
  title={K-planes: Explicit radiance fields in space, time, and appearance},
  author={Fridovich-Keil, Sara and Meanti, Giacomo and Warburg, Frederik Rahb{\ae}k and Recht, Benjamin and Kanazawa, Angjoo},
  booktitle={Proceedings of the IEEE/CVF conference on computer vision and pattern recognition},
  pages={12479--12488},
  year={2023}
}

@inproceedings{kang2026ilrm,
  title={ilrm: An iterative large 3d reconstruction model},
  author={Kang, Gyeongjin and Nam, Seungtae and Yang, Seungkwon and Sun, Xiangyu and Khamis, Sameh and Mohamed, Abdelrahman and Park, Eunbyung},
  booktitle={Proceedings of the IEEE/CVF Conference on Computer Vision and Pattern Recognition},
  pages={37332--37342},
  year={2026}
}

@inproceedings{nam2025generative,
  title={Generative densification: Learning to densify gaussians for high-fidelity generalizable 3d reconstruction},
  author={Nam, Seungtae and Sun, Xiangyu and Kang, Gyeongjin and Lee, Younggeun and Oh, Seungjun and Park, Eunbyung},
  booktitle={Proceedings of the Computer Vision and Pattern Recognition Conference},
  pages={26683--26693},
  year={2025}
}

@inproceedings{kang2026multi,
  title={Multi-view pyramid transformer: Look coarser to see broader},
  author={Kang, Gyeongjin and Yang, Seungkwon and Nam, Seungtae and Lee, Younggeun and Kim, Jungwoo and Park, Eunbyung},
  booktitle={Proceedings of the IEEE/CVF Conference on Computer Vision and Pattern Recognition},
  pages={37380--37390},
  year={2026}
}

@article{jin2024lvsm,
  title={Lvsm: A large view synthesis model with minimal 3d inductive bias},
  author={Jin, Haian and Jiang, Hanwen and Tan, Hao and Zhang, Kai and Bi, Sai and Zhang, Tianyuan and Luan, Fujun and Snavely, Noah and Xu, Zexiang},
  journal={arXiv preprint arXiv:2410.17242},
  year={2024}
}

@article{flynn2024quark,
  title={Quark: Real-time, high-resolution, and general neural view synthesis},
  author={Flynn, John and Broxton, Michael and Murmann, Lukas and Chai, Lucy and DuVall, Matthew and Godard, Cl{\'e}ment and Heal, Kathryn and Kaza, Srinivas and Lombardi, Stephen and Luo, Xuan and others},
  journal={ACM Transactions on Graphics (TOG)},
  volume={43},
  number={6},
  pages={1--20},
  year={2024},
  publisher={ACM New York, NY, USA}
}

@inproceedings{imtiaz2025lvt,
  title={LVT: Large-Scale Scene Reconstruction via Local View Transformers},
  author={Imtiaz, Tooba and Chai, Lucy and Heal, Kathryn and Luo, Xuan and Park, Jungyeon and Dy, Jennifer and Flynn, John},
  booktitle={Proceedings of the SIGGRAPH Asia 2025 Conference Papers},
  pages={1--12},
  year={2025}
}

@inproceedings{darcet2024vision,
  title={Vision transformers need registers},
  author={Darcet, Timoth{\'e}e and Oquab, Maxime and Mairal, Julien and Bojanowski, Piotr},
  booktitle={International conference on learning representations},
  volume={2024},
  pages={2632--2652},
  year={2024}
}

@misc{oquab2023dinov2,
  title={DINOv2: Learning Robust Visual Features without Supervision},
  author={Oquab, Maxime and Darcet, Timothée and Moutakanni, Theo and Vo, Huy V. and Szafraniec, Marc and Khalidov, Vasil and Fernandez, Pierre and Haziza, Daniel and Massa, Francisco and El-Nouby, Alaaeldin and Howes, Russell and Huang, Po-Yao and Xu, Hu and Sharma, Vasu and Li, Shang-Wen and Galuba, Wojciech and Rabbat, Mike and Assran, Mido and Ballas, Nicolas and Synnaeve, Gabriel and Misra, Ishan and Jegou, Herve and Mairal, Julien and Labatut, Patrick and Joulin, Armand and Bojanowski, Piotr},
  journal={arXiv:2304.07193},
  year={2023}
}

@article{simeoni2025dinov3,
  title={Dinov3},
  author={Sim{\'e}oni, Oriane and Vo, Huy V and Seitzer, Maximilian and Baldassarre, Federico and Oquab, Maxime and Jose, Cijo and Khalidov, Vasil and Szafraniec, Marc and Yi, Seungeun and Ramamonjisoa, Micha{\"e}l and others},
  journal={arXiv preprint arXiv:2508.10104},
  year={2025}
}

@article{zhou2018stereo,
  title={Stereo magnification: Learning view synthesis using multiplane images},
  author={Zhou, Tinghui and Tucker, Richard and Flynn, John and Fyffe, Graham and Snavely, Noah},
  journal={arXiv preprint arXiv:1805.09817},
  year={2018}
}

@inproceedings{ling2024dl3dv,
  title={Dl3dv-10k: A large-scale scene dataset for deep learning-based 3d vision},
  author={Ling, Lu and Sheng, Yichen and Tu, Zhi and Zhao, Wentian and Xin, Cheng and Wan, Kun and Yu, Lantao and Guo, Qianyu and Yu, Zixun and Lu, Yawen and others},
  booktitle={Proceedings of the IEEE/CVF Conference on Computer Vision and Pattern Recognition},
  pages={22160--22169},
  year={2024}
}

@inproceedings{jiang2024leap,
  title={Leap: Liberate sparse-view 3d modeling from camera poses},
  author={Jiang, Hanwen and Jiang, Zhenyu and Zhao, Yue and Huang, Qixing},
  booktitle={International Conference on Learning Representations},
  volume={2024},
  pages={12648--12667},
  year={2024}
}

@inproceedings{zhang2018lpips,
  title={The unreasonable effectiveness of deep features as a perceptual metric},
  author={Zhang, Richard and Isola, Phillip and Efros, Alexei A and Shechtman, Eli and Wang, Oliver},
  booktitle={Proceedings of the IEEE conference on computer vision and pattern recognition},
  pages={586--595},
  year={2018}
}

@ARTICLE{ssim,
  author={Zhou Wang and Bovik, A.C. and Sheikh, H.R. and Simoncelli, E.P.},
  journal={IEEE Transactions on Image Processing}, 
  title={Image quality assessment: from error visibility to structural similarity}, 
  year={2004},
  volume={13},
  number={4},
  pages={600-612},
  doi={10.1109/TIP.2003.819861}}

@inproceedings{ziwen2025long,
  title={Long-lrm: Long-sequence large reconstruction model for wide-coverage gaussian splats},
  author={Ziwen, Chen and Tan, Hao and Zhang, Kai and Bi, Sai and Luan, Fujun and Hong, Yicong and Fuxin, Li and Xu, Zexiang},
  booktitle={Proceedings of the IEEE/CVF International Conference on Computer Vision},
  pages={4349--4359},
  year={2025}
}

@book{hartley2003multiple,
  title={Multiple view geometry in computer vision},
  author={Hartley, Richard and Zisserman, Andrew},
  year={2003},
  publisher={Cambridge university press}
}

@inproceedings{yeshwanth2023scannet++,
  title={Scannet++: A high-fidelity dataset of 3d indoor scenes},
  author={Yeshwanth, Chandan and Liu, Yueh-Cheng and Nie{\ss}ner, Matthias and Dai, Angela},
  booktitle={Proceedings of the IEEE/CVF International Conference on Computer Vision},
  pages={12--22},
  year={2023}
}

@inproceedings{sun2026uni3r,
  title={Uni3r: Unified 3d reconstruction and semantic understanding via generalizable gaussian splatting from unposed multi-view images},
  author={Sun, Xiangyu and Jiang, Haoyi and Liu, Liu and Nam, Seungtae and Kang, Gyeongjin and Wang, Xinjie and Sui, Wei and Su, Zhizhong and Liu, Wenyu and Wang, Xinggang and others},
  booktitle={Proceedings of the IEEE/CVF Conference on Computer Vision and Pattern Recognition},
  pages={33280--33290},
  year={2026}
}

@article{loshchilov2017decoupled,
  title={Decoupled weight decay regularization},
  author={Loshchilov, Ilya and Hutter, Frank},
  journal={arXiv preprint arXiv:1711.05101},
  year={2017}
}

@article{kingma2014adam,
  title={Adam: A method for stochastic optimization},
  author={Kingma, Diederik P and Ba, Jimmy},
  journal={arXiv preprint arXiv:1412.6980},
  year={2014}
}

@article{barron2022mipnerf360,
    title={Mip-NeRF 360: Unbounded Anti-Aliased Neural Radiance Fields},
    author={Jonathan T. Barron and Ben Mildenhall and 
            Dor Verbin and Pratul P. Srinivasan and Peter Hedman},
    journal={CVPR},
    year={2022}
}

@article{Knapitsch2017tnt,
    author    = {Arno Knapitsch and Jaesik Park and Qian-Yi Zhou and Vladlen Koltun},
    title     = {Tanks and Temples: Benchmarking Large-Scale Scene Reconstruction},
    journal   = {ACM Transactions on Graphics},
    volume    = {36},
    number    = {4},
    year      = {2017},
}

@inproceedings{zhang2025flare,
  title={Flare: Feed-forward geometry, appearance and camera estimation from uncalibrated sparse views},
  author={Zhang, Shangzhan and Wang, Jianyuan and Xu, Yinghao and Xue, Nan and Rupprecht, Christian and Zhou, Xiaowei and Shen, Yujun and Wetzstein, Gordon},
  booktitle={Proceedings of the Computer Vision and Pattern Recognition Conference},
  pages={21936--21947},
  year={2025}
}

@inproceedings{reizenstein21co3d,
	Author = {Reizenstein, Jeremy and Shapovalov, Roman and Henzler, Philipp and Sbordone, Luca and Labatut, Patrick and Novotny, David},
	Booktitle = {International Conference on Computer Vision},
	Title = {Common Objects in 3D: Large-Scale Learning and Evaluation of Real-life 3D Category Reconstruction},
	Year = {2021},
}

@inproceedings{xia2024rgbd,
  title={Rgbd objects in the wild: Scaling real-world 3d object learning from rgb-d videos},
  author={Xia, Hongchi and Fu, Yang and Liu, Sifei and Wang, Xiaolong},
  booktitle={2024 IEEE/CVF Conference on Computer Vision and Pattern Recognition (CVPR)},
  pages={22378--22389},
  year={2024},
  organization={IEEE}
}

@inproceedings{jiang2025megasynth,
  title={Megasynth: Scaling up 3d scene reconstruction with synthesized data},
  author={Jiang, Hanwen and Xu, Zexiang and Xie, Desai and Chen, Ziwen and Jin, Haian and Luan, Fujun and Shu, Zhixin and Zhang, Kai and Bi, Sai and Sun, Xin and others},
  booktitle={2025 IEEE/CVF Conference on Computer Vision and Pattern Recognition (CVPR)},
  pages={16441--16452},
  year={2025},
  organization={IEEE}
}

@misc{jiang2026syn4d,
      title={Syn4D: A Multiview Synthetic 4D Dataset},
      author={Zeren Jiang and Yushi Lan and Yihang Luo and Yufan Deng and Zihang Lai and Edgar Sucar and Christian Rupprecht and Iro Laina and Diane Larlus and Chuanxia Zheng and Andrea Vedaldi},
      year={2026},
      eprint={2605.05207},
      archivePrefix={arXiv},
      primaryClass={cs.CV},
      url={https://arxiv.org/abs/2605.05207},
}

@inproceedings{yao2020blendedmvs,
      title={Blendedmvs: A large-scale dataset for generalized multi-view stereo networks},
      author={Yao, Yao and Luo, Zixin and Li, Shiwei and Zhang, Jingyang and Ren, Yufan and Zhou, Lei and Fang, Tian and Quan, Long},
      booktitle={Proceedings of the IEEE/CVF conference on computer vision and pattern recognition},
      pages={1790--1799},
      year={2020}
}

@inproceedings{Tosi2021CVPR,
      author = {Fabio Tosi and Yiyi Liao and Carolin Schmitt and Andreas Geiger},
      title = {SMD-Nets: Stereo Mixture Density Networks},
      booktitle = {Conference on Computer Vision and Pattern Recognition (CVPR)},
      year = {2021}
}

@inproceedings{DeepMVS,
      author       = "Huang, Po-Han and Matzen, Kevin and Kopf, Johannes and Ahuja, Narendra and Huang, Jia-Bin",
      title        = "DeepMVS: Learning Multi-View Stereopsis",
      booktitle    = "IEEE Conference on Computer Vision and Pattern Recognition (CVPR)",
      year         = "2018"
}

@inproceedings{schops2017multi,
  title={A multi-view stereo benchmark with high-resolution images and multi-camera videos},
  author={Schops, Thomas and Schonberger, Johannes L and Galliani, Silvano and Sattler, Torsten and Schindler, Konrad and Pollefeys, Marc and Geiger, Andreas},
  booktitle={Proceedings of the IEEE conference on computer vision and pattern recognition},
  pages={3260--3269},
  year={2017}
}

@inproceedings{solovev2023self,
  title={Self-improving multiplane-to-layer images for novel view synthesis},
  author={Solovev, Pavel and Khakhulin, Taras and Korzhenkov, Denis},
  booktitle={Proceedings of the IEEE/CVF Winter Conference on Applications of Computer Vision},
  pages={4309--4318},
  year={2023}
}
